\documentclass[11pt]{article}

\usepackage[final]{acl}

\usepackage{times}
\usepackage{latexsym}

\usepackage[T1]{fontenc}

\usepackage[utf8]{inputenc}

\usepackage{microtype}

\usepackage{inconsolata}

\usepackage{graphicx}
\usepackage{times}  
\usepackage{helvet}  
\usepackage{courier}  
\usepackage{hyperref} 
\usepackage{graphicx} 
\usepackage{amsmath}
\usepackage{amssymb}
\usepackage{natbib}  
\usepackage{caption} 
\usepackage{pifont}       

\usepackage{array}
\newcolumntype{C}[1]{>{\centering\arraybackslash}m{#1}}
\usepackage{algorithm}
\usepackage{algorithmic}
\usepackage{multirow}
\usepackage{soul}
\usepackage{amsmath}
\usepackage{booktabs}
\usepackage[table]{xcolor}
\usepackage{makecell}
\usepackage{adjustbox}

\newcommand{\hcmark}{\ding{52}\rotatebox[origin=c]{-9.2}{\kern-0.7em\ding{55}}}

\newcommand{\eg}{\textit{e.g.}}

\newcommand{\etc}{\textit{etc}}

\usepackage{titletoc}

\definecolor{darkgreen}{rgb}{0.0, 0.8, 0.0}
\definecolor{backred}{RGB}{255, 190, 190}
\definecolor{backblue}{RGB}{210, 230, 250}
\definecolor{backgreen}{RGB}{137, 215, 188}
\definecolor{backyellow}{RGB}{251, 218, 195}

\definecolor{cvprblue}{rgb}{0.21,0.49,0.74}

\title{
\begin{minipage}{.05\textwidth}
\centering
\vspace{-1pt}
\includegraphics[width=\linewidth]{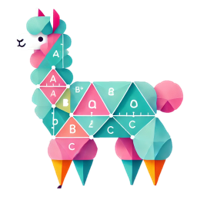} 
\end{minipage}
\texttt{MathFlow}: Enhancing the Perceptual Flow of MLLMs \\ for Visual Mathematical Problems}

\author{Shuhang Chen$^{1}$ \qquad Hangjie Yuan$^{1}$\thanks{Corresponding author.} \qquad Yunqiu Xu$^{1}$ \qquad Pengwei Liu$^{1}$   \\
\textbf{Tao Feng$^{3}$ \qquad  Jun Cen$^{1}$ \qquad Zeying Huang$^{2}$ \qquad   Yi Yang$^{1}$ }\\
$^{1}$Zhejiang University \qquad $^{2}$Intelligent Learning \qquad $^{3}$Tsinghua University\\
\texttt{\{sh.chen, hj.yuan, liupw, cenj, yangyics\}@zju.edu.cn} \\ 
\texttt{\{imyunqiuxu, fengtao.hi\}@gmail.com} \quad \texttt{jzxjeff@163.com}
}

\begin{document}
\maketitle
\begin{abstract}
Despite strong results on many tasks, multimodal large language models (MLLMs) still underperform on visual mathematical problem solving, especially in reliably perceiving and interpreting diagrams. Inspired by human problem-solving, we hypothesize that the ability to extract meaningful information from diagrams is pivotal, as it directly conditions subsequent inference.
Hence, we introduce FlowVerse, a comprehensive benchmark that provides a fine-grained evaluation of MLLMs’ perception and reasoning capabilities.
Our preliminary results on FlowVerse reveal that existing MLLMs exhibit substantial limitations when extracting essential information and reasoned properties from diagrams and performing complex reasoning based on these visual inputs. In response, we introduce MathFlow, a modular problem-solving pipeline that decouples perception and inference into distinct stages, thereby optimizing each independently. Given the perceptual limitations observed in current MLLMs, we trained MathFlow-P-7B as a dedicated perception model.
Experimental results indicate that MathFlow-P-7B yields substantial performance gains when integrated with various closed-source and open-source inference models. This demonstrates the effectiveness of the MathFlow pipeline and its compatibility with diverse inference frameworks.
Project page: \url{https://github.com/MathFlow-zju/MathFlow}.
\end{abstract}

\section{Introduction}

\begin{figure}[h]
    \centering
    \includegraphics[width=1\linewidth]{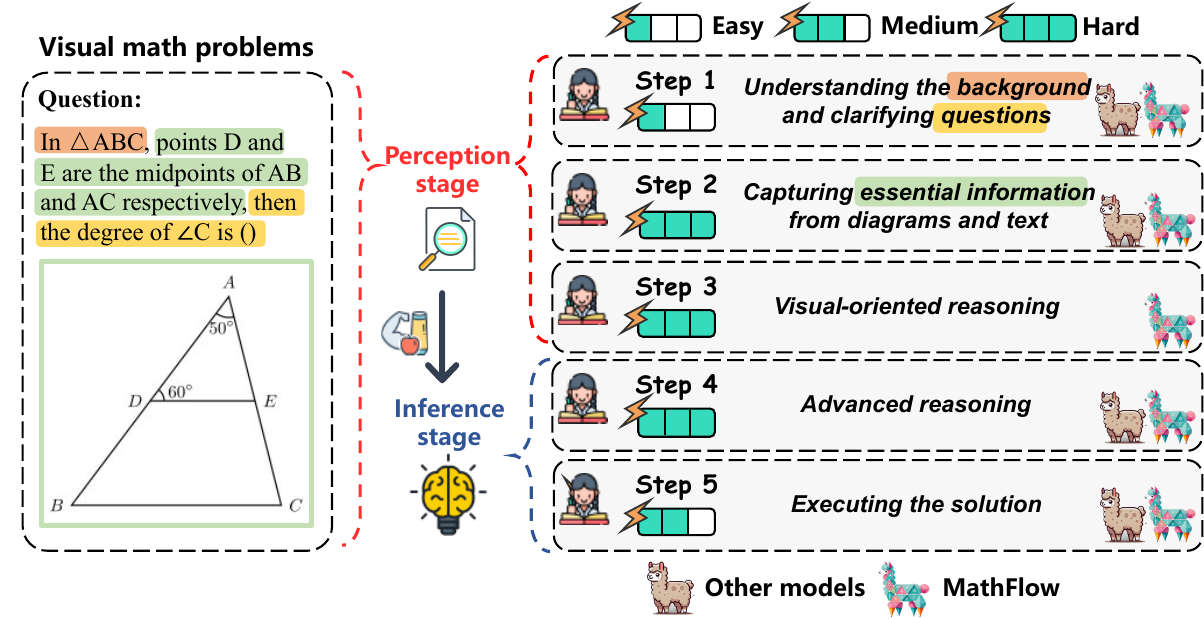}
   \vspace{-2em}
    \caption{\textbf{The Typical Process of Humans Solving Visual Mathematical
    Problems.} We can summarize two key capabilities observed in the typical human
    problem-solving process: perception and inference. The perception capability
    involves extracting relevant information from both visual and textual inputs,
    ensuring accurate reasoning, which inspired the development of FlowVerse and
    MathFlow.
    }
    \label{fig:motivation}
   \vspace{-0.5em}
\end{figure}

\begin{figure*}[h]
    \centering
    \includegraphics[width=1\linewidth]{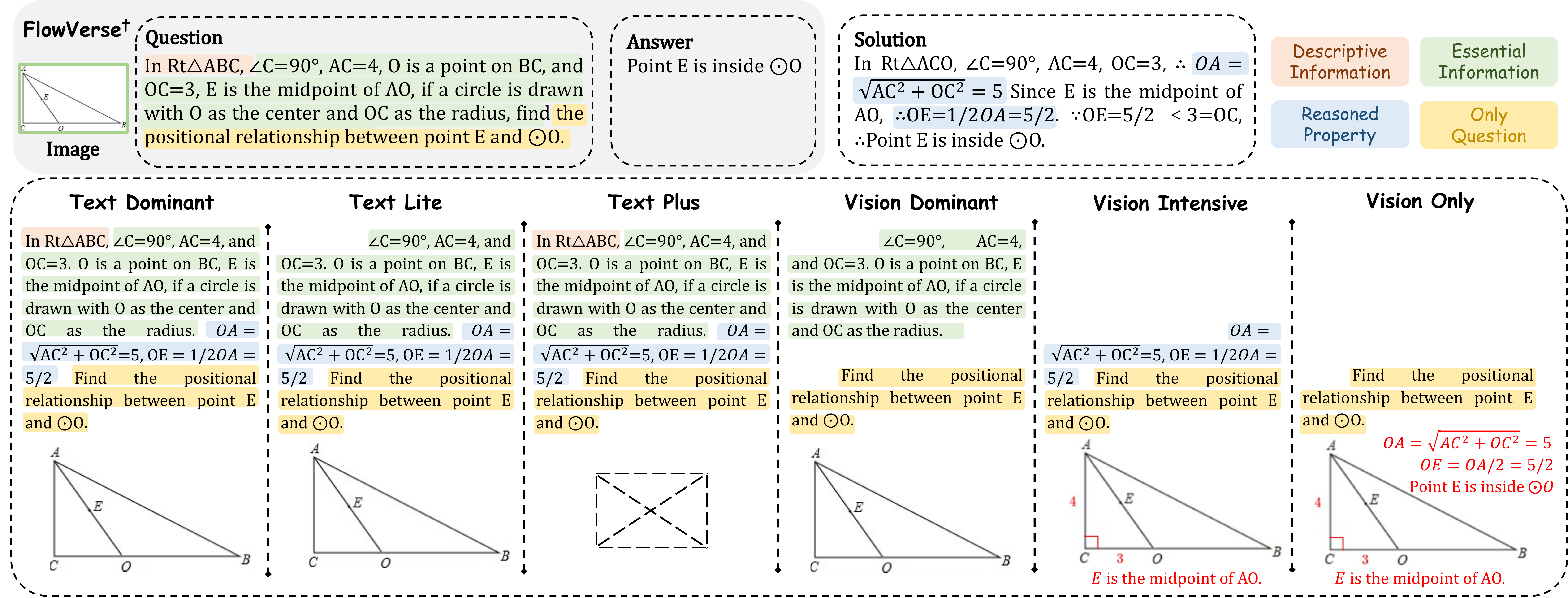}
   \vspace{-2em}
    \caption{ \textbf{Six Versions of Problems in FlowVerse.} FlowVerse begins
    by categorizing the original problem information into four distinct
    components: Descriptive Information (DI), Essential Information (EI), Only Question
    (OQ), and Reasoned Property (RP). The first three components are derived directly
    from the original problem statement, while RP is extracted from the solution
    and represents the inferences needed to solve the problem. In the Vision Centric
    version, we convert the EI into diagrams, while in the Vision Primary version,
    we convert both the EI and RP into diagrams. }
    \label{fig:enter-label}
   \vspace{-1em}
\end{figure*}

By enabling seamless interaction between visual data and natural language,
multimodal large language models (MLLMs)~\cite{gpt4o,bian2024make,feng2025zeroflow,feng2022overcoming,samora,xu2024gg,xu2025mc,li2026ma,xia2026echoes} are
excelling in various tasks, including captioning~\cite{Sph,jia2024mos2}, visual question
answering~\cite{qwen-vl} and visual dialogue~\cite{llava}. Despite their impressive
performance across diverse tasks, MLLMs have yet to fully demonstrate their
potential in visual mathematical problem-solving~\cite{unigeo,mathvista,chen2026cogflow}, particularly
in accurately perceiving and interpreting diagrams within these problems~\cite{mathverse}.

The typical process of humans for solving
visual mathematical problems consists of five sequential steps~\cite{krawec2014problem},
as illustrated in Fig.~\ref{fig:motivation}. We categorize them into two key stages:
\textit{1)} \textit{the perception stage} and \textit{2)} \textit{the inference
stage}. The perception stage focuses on extracting relevant information from both
visual and textual inputs that can be seamlessly integrated with the original problem
statement~\cite{polya1}. In contrast, the inference stage concentrates on mathematical reasoning.
Regarding the perception stage, 
existing MLLMs often struggle with mathematical images~\cite{blind,
geoqa} and extract unreliable information from visual contents, which
inevitably hampers the subsequent inference stage. Regarding the inference stage,
MLLMs often struggle to generate coherent solutions when essential perceptual information is limited or unreliable during complex problem solving~\cite{mathcheck, wemath}.
We propose that prior methods' limited capability to extract information during
the perception stage bounds the overall problem-solving performance. To validate
this assumption, we developed a new benchmark named FlowVerse. FlowVerse
categorizes all the information used during problem-solving into four parts~\cite{polya2}: \textit{Descriptive
Information (DI)}, which provides fundamental details describing the composition
of the problem; \textit{Essential Information (EI)}, which involves inferring
critical details from both diagrams and text; \textit{Only Question (OQ)}, representing
the specific question posed in the text; and \textit{Reasoned Property (RP)},
which refers to essential properties derived from diagrams and text during the
problem-solving process rather than explicitly being provided in the problem. We
design six problem variants using different combinations of information types,
as illustrated in Fig.~\ref{fig:enter-label}.

\begin{figure}[t]
    \centering
    \includegraphics[width=1\linewidth]{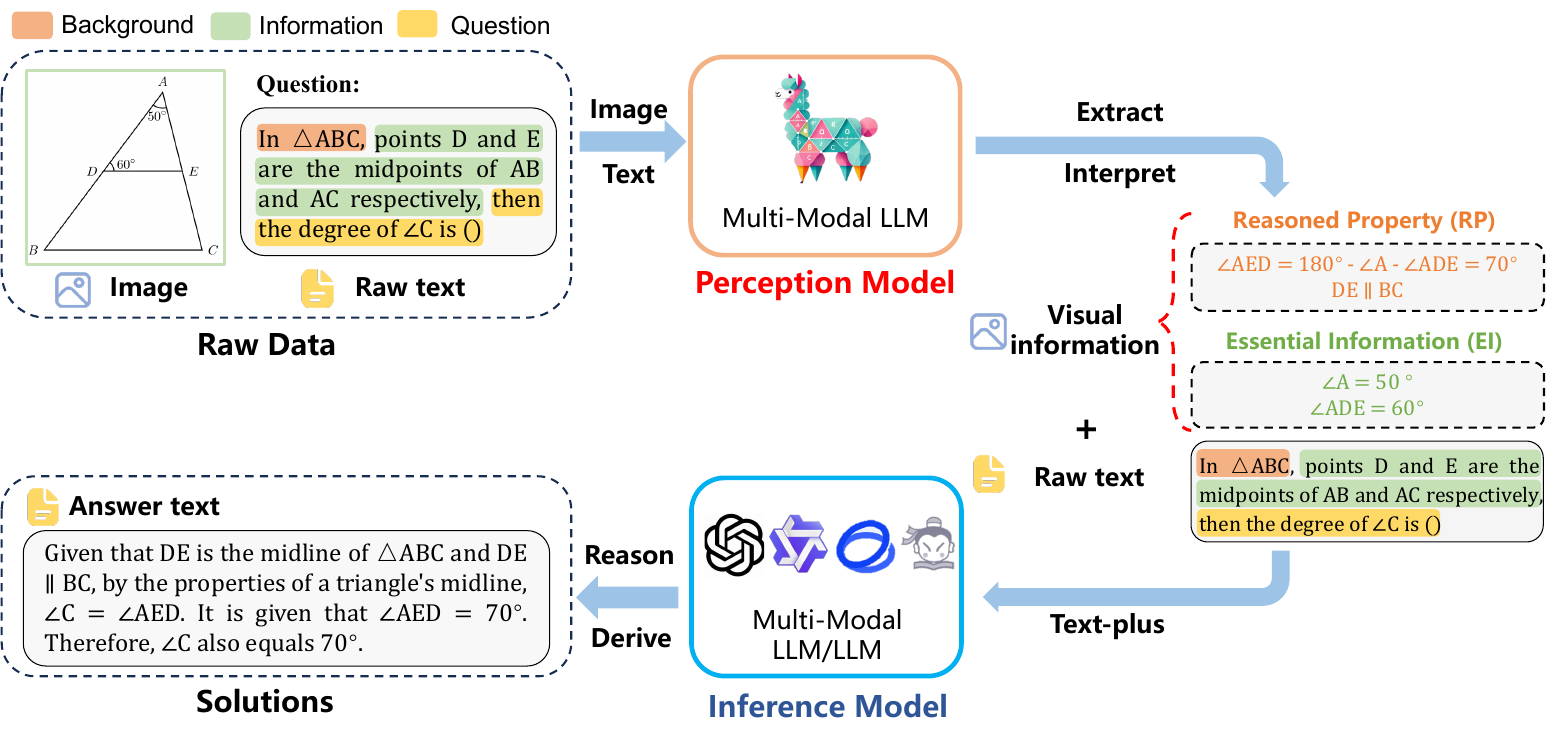}
   \vspace{-2em}
    \caption{\textbf{The Overview of MathFlow Pipeline.} To effectively train MLLMs for problem-solving, we decouple MLLMs into two sub-modules: the perception model and the inference model. 
    The perception model aims to extract and interpret visual information, converting it into a form that can be effectively processed.
    The inference model uses this extracted information, with the original question, to reason and derive solutions.}
    \label{fig:overview}
   \vspace{-0.5em}
\end{figure}

Our preliminary results on FlowVerse reveal that existing MLLMs struggle
significantly in extracting EI and RP from diagrams, as well as in carrying out complex
reasoning on given images. Motivated by these observations, we introduce a modular
problem-solving pipeline, MathFlow, which explicitly decouples the problem-solving
pipeline into two distinct stages: a perception stage and an inference stage.  Fig.~\ref{fig:overview} illustrates that EI is first extracted from diagrams, followed by preliminary
reasoning to derive RP in
the perception stage. They are then converted into a text-level representation,
further concatenated with the original problem's textual information to create a
unified, enriched input. The inference model can easily generate a reasonable solution
using the combined visual-derived and textual information.

Preliminary experiments on FlowVerse highlight the critical importance of robust
perception capabilities in deriving useful properties from visual mathematical diagrams.
However, a model as advanced as GPT-4V may exhibit deficiencies in this specific aspect~\cite{blind,feng2024more,ahn2024large,examining}.
To address this, we propose MathFlow-P-7B, a specialized model designed to
extract EI and RP information from visual diagrams, thereby facilitating visual mathematical
problem-solving. The training of MathFlow-P-7B involves two stages: the multi-task
pretraining stage and the supervised fine-tuning stage. The multi-task
pretraining stage includes two core tasks—captioning EI and RP. The supervised fine-tuning
stage further refines the model's perception capabilities for enhanced performance.
Experimental results indicate that MathFlow-P-7B enables superior performance
improvements when integrated with different inference models, underscoring the effectiveness
of the MathFlow pipeline.
Our contributions are summarized as follows:
\begin{itemize}

\item \textbf{FlowVerse benchmark:} We introduce a comprehensive benchmark specifically designed to evaluate the visual mathematical problem-solving capabilities of MLLMs across its meticulously crafted six problem versions.
\item \textbf{MathFlow pipeline:} We propose a modular pipeline that decouples the visual mathematical problem-solving process into perception and inference, not only enhancing the model's ability to extract and reason with multimodal information effectively, but also empowering LLMs to handle visual math problems.
\item \textbf{MathFlow-P-7B model:} We develop a perceptive mathematics model, an MLLM specifically optimized for visual mathematical problems, demonstrating state-of-the-art performance across various benchmarks.
\end{itemize}

\begin{table*}[!t]
\small
\centering
\caption{\textbf{Mathematical Evaluation on Six Problem Versions in FlowVerse}. 
DI, EI, RP, OQ refer to the {textual} or {visual} Descriptive Information, Reasoned Property, Essential Information, Only Question, respectively. 
The Text Plus Version does not involve image input. 
``CoT-E'' or ``Acc'' denotes whether to employ the FlowVerse-CoT-E strategy or not. 
The highest accuracy for each group of MLLMs is marked in \textbf{bold}. 
\textit{The full table is provided in Appendix~\S\ref{More_Detailed_Experiment_Results}.}}
\vspace{-1em}
\begin{adjustbox}{width=\linewidth}
\begin{tabular}{l|C{0.9cm}C{0.9cm}|C{0.9cm}C{0.9cm}|C{0.9cm}C{0.9cm}|C{0.9cm}C{0.9cm}|C{0.9cm}C{0.9cm}|C{0.9cm}C{0.9cm}|C{0.9cm}C{0.9cm}}
\toprule
\multirow{4}*{\makecell*[l]{\large \textbf{Model}}}    
& \multicolumn{2}{c|}{\textbf{All}} 
& \multicolumn{2}{c|}{\textbf{Text Centric}} 
& \multicolumn{2}{c|}{\textbf{Text Limited}}
& \multicolumn{2}{c|}{\textbf{Text Plus}}
& \multicolumn{2}{c|}{\textbf{Vision Dense}}
& \multicolumn{2}{c|}{\textbf{Vision Centric}}
& \multicolumn{2}{c}{\textbf{Vision Primary}} \\
~ & \multicolumn{2}{c|}{~} 
& \multicolumn{2}{c|}{DI+RP+EI+OQ} 
& \multicolumn{2}{c|}{RP+EI+OQ} 
& \multicolumn{2}{c|}{DI+RP+EI+OQ} 
& \multicolumn{2}{c|}{EI+OQ} 
& \multicolumn{2}{c|}{RP+EI+OQ} 
& \multicolumn{2}{c}{RP+EI+OQ} \\
\cmidrule{2-15}
~ & CoT-E & Acc & CoT-E & Acc & CoT-E & Acc & CoT-E & Acc & CoT-E & Acc & CoT-E & Acc & CoT-E & Acc \\
\midrule
\multicolumn{15}{c}{\textit{Open-source MLLMs}}\\
\midrule
InfiMM-Math-7B &36.5&28.8&43.8&38.1&40.6&36.7&46.1&40.1&28.8&15.4&39.6&30.3&26.1&23.2\\
InternVL2.5-8B &44.7&41.0&49.2&41.3&40.5&38.4&49.6&42.7&38.4&20.2&41.0&35.9&35.8&33.9\\
Qwen2.5-VL-7B  &53.8&42.2&60.1&52.8&58.9&51.3&62.0&55.0&45.0&31.0&50.8&46.3&48.1&45.3\\
VLM-R1-7B$^{\dagger}$ &50.7&41.2&59.0&54.2&57.9&49.8&65.5&58.9&36.2&24.5&46.1&37.8&30.6&26.1\\
Qwen2-VL-72B   &52.3&48.6&59.4&47.3&54.3&45.7&63.7&50.0&40.8&25.3&50.9&42.1&47.6&37.0\\
InternVL2.5-78B&54.7&50.1&66.1&62.7&64.1&60.3&67.8&64.7&48.7&34.3&63.0&58.8&59.6&57.7\\
\rowcolor{backblue!75}
MathFlow$^{\star}_{\text{Qwen2.5\!-\!VL\!-\!7B}}$
&57.0&46.0&62.0&53.8&60.8&52.2&64.2&56.0&49.0&39.1&54.5&51.6&52.0&51.5\\
\midrule
\multicolumn{15}{c}{\textit{Math-specialized MLLMs}}\\
\midrule
Math\!-\!LLaVA\!-\!13B &38.0&29.9&45.1&39.3&44.4&37.4&--&--&36.2&18.6&41.7&35.9&37.0&34.2\\
MultiMath-7B &44.0&34.2&50.6&44.8&49.9&42.9&--&--&41.7&22.1&47.2&40.4&39.7&38.8\\
SVE-Math-Qwen2.5-7B &46.0&39.4&53.1&47.3&53.4&45.8&--&--&44.2&28.6&48.9&44.2&45.8&42.0\\
\midrule
\multicolumn{15}{c}{\textit{Closed-source MLLMs}}\\
\midrule
Qwen-VL-Max &43.0&36.3&49.8&42.1&46.7&38.3&53.9&51.0&38.6&15.2&42.7&33.2&29.6&27.8\\
GPT-4o-mini  &51.3&44.5&58.7&54.8&58.2&53.2&59.6&55.2&41.1&26.0&57.4&50.1&49.7&47.6\\
Claude-3.5-Sonnet &56.9&49.6&60.8&52.6&58.7&50.3&64.0&58.3&45.0&25.4&56.5&48.0&48.1&45.2\\
GPT-4o       &55.1&47.8&61.0&56.8&58.7&54.4&62.2&58.2&45.2&30.0&58.6&52.6&54.1&51.0\\
GPT-4V       &56.2&53.4&69.1&57.1&65.0&55.0&72.0&61.4&48.1&30.3&61.8&46.3&42.0&36.7\\
Gemini-2.5-pro&62.0&55.3&68.3&61.9&66.1&60.8&68.9&64.1&52.1&37.1&65.7&57.9&57.0&54.6\\
GPT-5        &  {65.8}&  {60.1}&  {74.3}&  {68.1}&  {73.5}&  {66.7}&  {77.0}&  {69.2}&  {53.8}&  {44.7}&  {67.1}&61.7&60.3&  {57.5}\\
\rowcolor{backblue!75}
MathFlow$^{\star}_{\text{GPT\!-\!5}}$
&\textbf{66.5}&\textbf{61.8}&\textbf{74.6}&\textbf{68.5}&\textbf{73.8}&\textbf{67.2}&\textbf{77.0}&\textbf{69.3}&\textbf{58.2}&\textbf{54.1}&\textbf{70.2}&\textbf{66.7}&\textbf{69.4}&\textbf{66.2}\\
\bottomrule
\end{tabular}
\end{adjustbox}
\label{tab:flowverse_benchmark}
\vspace{-0.5em}
\end{table*}

\section{Related Work}
\noindent
\textbf{Visual Mathematical Reasoning with MLLMs.}
Solving visual mathematical problems (\eg, geometry diagrams, algebraic plots, and \etc) requires both strong reasoning ability and accurate interpretation of visual primitives and symbolic content~\citep{yan2024survey,wemath2}.
Most previous works are dedicated to improving the reasoning process, including chain-of-thought strategies~\citep{xu2025llavacotletvisionlanguage,r-cot}, tool-aided reasoning~\citep{alphageometry,pot}, test time scaling~\citep{visuothink,vstar}, and reinforcement learning~\citep{Skyworkr1v2,vlm-r3}.
Several recent works~\citep{DVLR,vcar,wei2025slowperceptionletsperceive} suggest that one of the major bottlenecks in visual mathematical reasoning is inaccurate visual comprehension.

\vspace{0.3em}\noindent\textbf{Mathematics Evaluation Benchmarks.}
The use of LLMs and MLLMs for solving visual mathematical problems has been extensively explored in several studies~\cite{tsp,wang2024exploring,mathglm,multimath}. 
To better assess the progress of LLMs and MLLMs in mathematical solving and to drive further improvement, a number of mathematics-specific evaluation benchmarks have been introduced~\cite{liu2024finemath,satpute2024can,zhang2024mario}.
However, many mathematical problems include diagrams, which has led researchers to increasingly focus on developing MLLMs that can effectively handle both textual and visual content~\cite{g-llava,eagle}.
To provide a more comprehensive and in-depth evaluation of MLLMs, MathVerse~\cite{mathverse} builds upon MathVista, focusing on eliminating textual redundancy to ensure genuine interpretation of visual diagrams rather than reliance on textual shortcuts. It also introduces a Chain-of-Thought (CoT) evaluation strategy~\cite{cot,wu2024chain} for a detailed assessment of intermediate reasoning steps. However, our analysis reveals that these benchmark datasets still have limitations.

\section{FlowVerse Benchmark}
\label{subsec:flowverse}

To validate MLLMs' capabilities regarding perception and inference, we first propose FlowVerse, which categorizes problem information into several key components.

\vspace{0.3em}\noindent\textbf{Dataset Composition.} 
FlowVerse comprises 2,000 visual mathematical
problems collected from real exam questions in both Chinese and English, 
resulting in over 12,000 test samples. \textit{Detailed statistics
for the dataset composition are presented in Tab.~\ref{supp-t5} in Appendix~\S\ref{Detailed_Statistics_of_FlowVerse}.}
To guarantee the dataset’s quality and precision, we conducted a thorough review
to verify the accuracy of the answers and analyses, as well as to ensure consistency
between the questions and their corresponding diagrams. Notably, to avoid any confusion,
we named the initially collected dataset\textbf{ FlowVerse$^{\dagger}$}, while the
subsequent dataset containing six versions of the problem is called \textbf{FlowVerse}.

Furthermore, to achieve a comprehensive evaluation, this meticulously curated
dataset encompasses four key components of information flow: \ding{182} \textbf{\textit{Descriptive Information}} (DI) , \ding{183} \textbf{\textit{Essential Information}} (EI), \ding{184} \textbf{\textit{Reasoned Property}} (RP), and \ding{185} \textbf{\textit{Only Question}} (OQ).
Each of these components is carefully crafted to provide distinct insights into the
factors that influence the perception capabilities of MLLMs: DI provides fundamental
        details about the composition of figures, referring directly to the observable
        and explicitly depicted elements within a diagram, such as geometric
        shapes or intersection points in functions. These descriptions help establish
        context and frame the problem.
EI refers to the critical
        details required for problem-solving, such as specific values or relationships
        between geometric elements (\eg, $\angle A = 45^{\circ}$, $AD \perp BC$).
        FlowVerse incorporates information directly from visual diagrams as part
        of EI, recognizing that in visual math problems, much of the essential information
        comes from the diagrams. Thus, accurately extracting EI from multiple modalities
        is crucial for MLLMs to solve problems effectively.
RP represents information
        inferred through higher-level visual abstraction and reasoning combined
        with relevant mathematical knowledge. Unlike EI, RP requires
        understanding beyond simple visual perception, as diagrams may not explicitly
        convey relationships between geometric elements. In FlowVerse, RP is often
        not directly present in the original problem but requires additional human
        annotation. These annotations serve as guiding tips to assist solvers, enhancing
        accuracy and enabling a more comprehensive and fine-grained evaluation of
        MLLMs' capabilities. Thus, RP in FlowVerse is neither a fixed output nor a mere hint, but a controlled variable that allows us to independently examine (1) whether a model can generate RP, (2) whether it can utilize provided RP, and (3) how the modality of RP influences performance. This diagnostic design enables interpretable separation of visual perception, RP generation, and RP utilization, ensuring that no conceptual conflict arises.
OQ refers to the specific question
        posed in the text.
        Though a small part of the problem, it determines what needs to be solved, without which the solution process cannot proceed.

Based on the four categories, expert annotators systematically remove different types
of information within questions and progressively incorporate critical elements into
the diagrams. As illustrated in Fig~\ref{fig:enter-label}, we generate six versions
of each problem based on four key components of information, resulting in 12,000
test instances. With this curated problem set, we provide a comprehensive
evaluation of the visual perception capabilities of MLLMs and assess whether
these abilities can effectively support multimodal mathematical reasoning. The details
of each problem version are:
   \ding{172}  \textbf{\textit{Text Centric Version}} retains all content, including DI,
        RP, EI, and OQ. Notably, in this version, we manually convert all RP and
        EI components into textual form to effectively analyze their impact on the
        final reasoning process across different modalities. 
  \ding{173}  \textbf{\textit{Text Limited Version}} removes the DI from the Text
        Centric Version while retaining the other components.
 \ding{174}   \textbf{\textit{Text Plus Version}} excludes the image entirely from
        the Text Centric Version, reducing the input from a multimodal to a purely
        text-based format.
 \ding{175} \textbf{\textit{Vision Dense Version}} removes the RP from the Text
        Limited Version.
   \ding{176} \textbf{\textit{Vision Centric Version}} converts the EI  from text to image, starting with the Text Limited Version,
        thereby making it more visually focused.
   \ding{177} \textbf{\textit{Vision Primary Version}} further converts the Reasoned
        Property from text to image, based on the Vision Centric Version,
        resulting in an entirely visual input for analysis.

Finally,  the purpose of incorporating RP among these variants is not to provide the model with direct reasoning hints, but rather to decouple and evaluate how different modalities contribute to problem-solving performance. RP serves as diagnostic information rather than a substitute for the full reasoning chain. 
In certain variants (\eg, Text Centric), RP is supplied to assess a model’s pure inference ability when the necessary intermediate properties are already known. 
In other variants (\eg, Vision Dense), RP is removed or presented visually, requiring the model to infer these properties from the diagram itself. 
Comparing variants enables systematic decomposition of distinct capabilities.

\begin{figure}[t]
    \centering
    \includegraphics[width=1\linewidth]{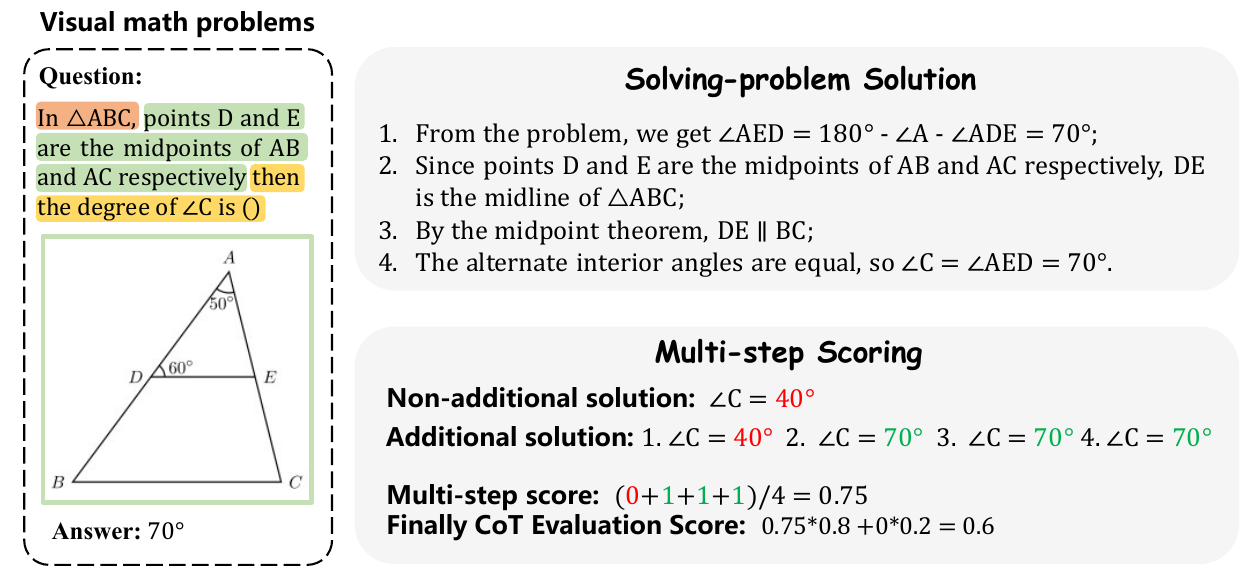}
   \vspace{-2em}
    \caption{\textbf{The FlowVerse-CoT-E Strategy.}}
    \label{fig:cot-e}
   \vspace{-0.5em}
\end{figure}

\vspace{0.3em}\noindent\textbf{FlowVerse-CoT-E: Evaluation Strategy.} As previously mentioned, several studies have evaluated the abilities of MLLMs from a Chain-of-Thought (CoT) perspective~\cite{cot}. However, these approaches typically rely on GPT or human intervention to decompose the model’s response into sequential steps and subsequently evaluate the correctness of each step. The variability of intermediate steps complicates this process, as GPT’s interpretation and decomposition of questions can introduce errors (termed MathVerse-CoT-E), leading to significant noise and undermining the accuracy and consistency of subsequent evaluations~\cite{qwen-vl,internlm}. To address these challenges, we propose a \textbf{robust CoT-based evaluation strategy}, termed FlowVerse-CoT-E, which sequentially integrates components of an authoritative problem-solving solution into the prompts. Specifically, experts first develop an authoritative problem-solving method for each question in FlowVerse and then decompose it into several solution steps. These steps are progressively incorporated into the prompts provided to the MLLMs for reasoning. As illustrated in Fig.~\ref{fig:cot-e}, the results generated by MLLMs under different prompts are aggregated using a weighted approach to derive the final outcome. Notably, to ensure the robustness of the evaluation, we provide multiple solutions for questions that admit more than one valid approach.
The scoring process is formulated as follows:
\vspace{-0.5em}
\begin{equation}
    \text{Score}_\text{final}= \alpha(\frac{1}{N}\sum_{i=1}^{N}\text{Score}_{i}) + (1-
    \alpha)\text{Score}_{0},
\vspace{-0.5em}
\end{equation}
where $\alpha$ defines as a balancing factor between intermediate reasoning steps, setting $\alpha = 0.8$ by default to highlight the importance of CoT reasoning. $\text{Score}_i$ represents the MLLM’s score up to step $i$, while $\text{Score}_0$ reflects its score based solely on the final answer, without intermediate steps.

\vspace{0.3em}\noindent
\textbf{Key Findings on FlowVerse.} 
As reported in Tab.~\ref{tab:flowverse_benchmark}, our preliminary experiments on FlowVerse benchmark highlight the critical importance of robust perception capabilities in deriving useful properties from visual mathematical diagrams.
\textbf{\textit{Please refer to \S\ref{sec:flowverse_results} for more detailed discussions.}}

\section{MathFlow: A Modular Visual Mathematical Problem-Solving Pipeline}
\label{sec:mathflow}

Motivated by the key findings illustrated above, we aim to enhance the extraction of accurate textual EI and RP. Hence, we introduce a modular problem-solving pipeline, MathFlow.
The MathFlow pipeline consists of two stages to alleviate the limitations of MLLMs in visual mathematical problem-solving. In the perception stage, the model extracts critical information from visual data and converts it into a text representation, integrating it with the original problem. This enriched input is then passed to the inference stage.

\begin{table*}[t] 
    \small
    \centering
    \caption{\textbf{Mathematical Evaluation on Six Problem Versions in
    MathVerse's \textit{testmini} Set.} The ``All" score is calculated without
    including the average of the Text Only version. ``CoT-E" or ``Acc" indicates
    the use of the proposed CoT evaluation strategy or not. $^{\star}$ denotes
    that MathFlow-P-7B functions as the perception model.
    The highest accuracy for closed-source and open-source MLLMs is marked in
    \textbf{blod} and \ul{uline} respectively.}
   \vspace{-1em}
    \begin{adjustbox}
        {width=\linewidth}
        \begin{tabular}{l|C{0.9cm}C{0.9cm}|C{0.9cm}C{0.9cm}|C{0.9cm}C{0.9cm}|C{0.9cm}C{0.9cm}|C{0.9cm}C{0.9cm}|C{0.9cm}C{0.9cm}|C{0.9cm}C{0.9cm}}
            \toprule \multirow{3}*{\makecell*[l]{\large \textbf{Model}}} & \multicolumn{2}{c|}{\makecell*[c]{\textbf{All}}} & \multicolumn{2}{c|}{\makecell*[c]{\shortstack{\vspace*{0.1pt}\\\textbf{Text}\\\vspace*{0.2pt}\\\textbf{Dominant}}}} & \multicolumn{2}{c|}{\makecell*[c]{\shortstack{\vspace*{0.1pt}\\\textbf{Text}\\\vspace*{0.2pt}\\\textbf{Lite}}}} & \multicolumn{2}{c|}{\makecell*[c]{\shortstack{\vspace*{0.1pt}\\\textbf{Text}\\\vspace*{0.2pt}\\\textbf{Only}}}} & \multicolumn{2}{c|}{\makecell*[c]{\shortstack{\vspace*{0.1pt}\\\ \textbf{Vision}\ \ \\\vspace*{0.2pt}\\\textbf{Intensive}}}} & \multicolumn{2}{c|}{\makecell*[c]{\shortstack{\vspace*{0.1pt}\\\ \textbf{Vision}\ \ \\\vspace*{0.2pt}\\\textbf{Dominant}}}} & \multicolumn{2}{c}{\makecell*[c]{\shortstack{\vspace*{0.1pt}\\\ \textbf{Vision}\ \ \\\vspace*{0.2pt}\\\textbf{Only}}}} \\
            \cmidrule{2-15} ~                                   & CoT-E                                   & \ Acc\                                                                                            & CoT-E                                                                                         & Acc                                                                                           & CoT-E                                                                                                      & Acc                                                                                                       & CoT-E                                                                                               & Acc                          & CoT-E                        & Acc                          & CoT-E                        & Acc                          & CoT-E                        & Acc                          \\
            \midrule Qwen-VL-Plus              & 21.3                                    & 11.8                                                                                              & 26.0                                                                                          & 15.7                                                                                          & 21.2                                                                                                       & 11.1                                                                                                      & 25.2                                                                                                & 14.5                         & 18.5                         & 9.0                          & 19.1                         & 13.0                         & 21.8                         & 10.0                         \\
            Gemini-Pro                            & 35.3                                    & 23.5                                                                                              & 39.8                                                                                          & 26.3                                                                                          & 34.7                                                                                                       & 23.5                                                                                                      & 44.5                                                                                                & 27.3                         & 32.0                         & 23.0                         & 36.8                         & 22.3                         & 33.3                         & 22.2                         \\
            Qwen-VL-Max                          & 37.2                                    & 25.3                                                                                              & 42.8                                                                                          & 30.7                                                                                          & 37.7                                                                                                       & 26.1                                                                                                      & 47.9                                                                                                & 28.9                         & 33.6                         & 24.1                         & 35.9                         & 24.1                         & 35.9                         & 21.4                         \\
            GPT-4V                                & 54.4                                    & 39.4                                                                                              & 63.1                                                                                          & 54.7                                                                                          & 56.6                                                                                                       & 41.4                                                                                                      & 60.3                                                                                                & 48.7                         & 51.4                         & 34.9                         & 50.8                         & 34.4                         & 50.3                         & 31.6                         \\ \rowcolor{backblue!75}
            MathFlow$^{\star}$$_{\text{GPT-4V}}$              & \textbf{56.7}             & \textbf{43.8}                                                                       & \textbf{65.2}                                                                   & \textbf{51.1}                                                                   & \textbf{58.9}                                                                                & \textbf{46.4}                                                                               & \textbf{62.1}                                                                         & \textbf{48.5}  & \textbf{53.7}  & \textbf{40.3}  & \textbf{52.1 } & \textbf{37.4}  & \textbf{52.5}  & \textbf{39.0}  \\
            \cmidrule{1-15} SPHINX-MoE            & 25.8                                    & {15.6}                                                                                            & 33.3                                                                                          & {22.2}                                                                                        & 21.9                                                                                                       & 16.4                                                                                                      & 40.7                                                                                                & {18.3}                       & 21.1                         & 14.8                         & 19.6                         & 12.6                         & 18.3                         & 9.1                          \\
            InternLM-XC2                        & 25.9                                    & 16.5                                                                                              & 36.9                                                                                          & 22.3                                                                                          & 28.3                                                                                                       & 17.0                                                                                                      & 42.5                                                                                                & 16.5                         & 20.1                         & 15.7                         & 24.4                         & 16.4                         & 19.8                         & 11.0                         \\
            MAVIS-7B                              & 27.5                                    & -                                                                                                 & 41.4                                                                                          & -                                                                                             & 29.1                                                                                                       & -                                                                                                         & 42.5                                                                                                & -                            & 27.4                         & -                            & 24.9                         & -                            & 14.6                         & -                            \\
            InfiMM-Math                           & 34.5                                    & -                                                                                                 & 46.7                                                                                          & -                                                                                             & 32.4                                                                                                       & -                                                                                                         & -                                                                                                   & -                            & {38.1}                       & -                            & 32.4                         & -                            & 15.8                         & -                            \\
            Qwen2-VL-72B                        & 38.9                                    & 28.5                                                                                              & 49.2                                                                                          & 35.4                                                                                          & 35.6                                                                                                       & 27.8                                                                                                      & 52.1                                                                                                & 39.6                         & 38.7                         & 26.4                         & 35.1                         & 28.5                         & 22.5                         & 13.2                         \\ \rowcolor{backblue!75}
            MathFlow$^{\star}$$_{\text{Qwen2-VL-72B}}$        & \ul{40.5}            & \ul{31.7}                                                                      & \ul{52.3}                                                                  & \ul{39.3}                                                                  & \ul{37.7}                                                                               & \ul{31.7}                                                                              & \ul{55.2}                                                                        & \ul{45.5} & \ul{40.5} & \ul{30.3} & \ul{37.2} & \ul{32.4} & \ul{24.6} & \ul{17.1} \\
            \bottomrule
        \end{tabular}
    \end{adjustbox}
    \label{tab:tab2}
   \vspace{-0.5em}
\end{table*}

A model's perception and inference capabilities are inherently distinct, with
strong perception being a prerequisite for accurate inference. Consequently, we
prioritize enhancing the model's perception abilities to enable precise
extraction and interpretation of visual information, thereby supporting more reliable
inference. Then, the inference stage can utilize other state-of-the-art MLLMs/LLMs,
allowing the best inference capabilities to complement the perception
improvements brought by MathFlow. This modular methodology ensures flexibility and
robustness in solving complex visual mathematical problems. For simplicity, we refer
to the model in the perception stage as ``perception model'' and the model in
the inference stage as ``inference model''.

Furthermore, MathFlow's training strategy is divided into two stages: the multi-task
pretraining stage and the supervised fine-tuning stage. Upon completion of training, we obtain a fine-tuned MLLM (MathFlow-P-7B).

\vspace{0.3em}\noindent
\textbf{Multi-Task Pretraining Stage.} The training tasks in this stage primarily
include the EI caption task and the visual-oriented reasoning task.

On the one hand, the EI caption task aims to train the perception model to
generate textual descriptions of essential elements directly from visual inputs.
For this, we fine-tune a pretrained Qwen2-VL-7B model~\cite{qwen2-vl} by
utilizing datasets like MAVIS~\cite{mavis} and Geo170k~\cite{geo170}, which contain
real or synthetic visual mathematics image-text pairs.

On the other hand, the visual-oriented reasoning task is designed to extract higher-level
abstractions and relationships, requiring the model to engage in abstract reasoning—such
as deducing relationships between geometric shapes or identifying intersection points
in a function. To support this task, we developed a training dataset called \textbf{\textit{MathFlow-RP}},
sampled from educational materials that include detailed problem-solving solutions.
\textit{\textbf{Details of MathFlow-RP are presented in Appendix~\S\ref{Details_of_MathFlow-RP}.}}
Specifically, we deconstructed each problem's corresponding solution into multiple
steps, using the previous step as context to guide the MLLM in predicting the
next step of the solution.

Finally, we performed mixed training on both tasks, ensuring that the data ratio
between the two tasks was maintained at 3:1. During this stage, the LLM backbone
is frozen, and training is focused on the perceiver resampler and vision encoder
module to ensure that the visual components can effectively extract and process
the necessary information.

\vspace{0.3em}\noindent
\textbf{Supervised Fine-Tuning Stage.}
In this stage, our goal is to enhance the model's response quality and adapt it
more effectively to the current task context. To achieve this, we meticulously
developed a supervised fine-tuning dataset called \textbf{\textit{MathFlow-SFT}},
focusing on retaining only specific numerical values, relationships between geometric
elements, and essential information necessary for defining the solution space. {\textit{Details
of this process are provided in Appendix~\S\ref{Details_of_MathFlow-SFT}.} During this stage, we freeze the vision
encoder and concentrate on training the perceiver resampler and LLM backbone.
This approach aims to improve the model's ability to accurately interpret visual inputs and generate concise textual representations crucial for problem-solving.

\begin{table*}
   [!t] \small
   \centering
   \caption{\textbf{Comparison of model performances across various mathematical
   subjects on the MathVision dataset.} The highest accuracy for closed-source and
   open-source MLLMs is marked in \textbf{bold} and \ul{uline} respectively.}
   \vspace{-1em}
   \setlength{\tabcolsep}{2pt}
   \begin{adjustbox}
      {width=\linewidth}
      \begin{tabular}{l|C{0.9cm}C{0.9cm}C{0.9cm}C{0.9cm}C{0.9cm}C{0.9cm}C{0.9cm}C{0.9cm}C{0.9cm}C{0.9cm}C{0.9cm}C{0.9cm}C{0.9cm}C{0.9cm}C{0.9cm}C{0.9cm}C{0.9cm}C{0.9cm}}
         \toprule \textbf{Model}                               & \textbf{Overall}                       & \textbf{Alg}                          & \textbf{AnaG}                         & \textbf{Ari}                          & \textbf{CombG}                        & \textbf{Comb}                         & \textbf{Cnt}                          & \textbf{DescG}                        & \textbf{GrphT}                        & \textbf{Log}                          & \textbf{Angle}                        & \textbf{Area}                         & \textbf{Len}                          & \textbf{SolG}                         & \textbf{Stat}                         & \textbf{Topo}                         & \textbf{TransG}                       \\
         \midrule Qwen-VL-Plus                        & 10.72                         & 11.3                         & 17.9                         & 14.3                         & 12.7                         & 4.8                          & 10.5                         & 15.4                         & 8.9                          & 14.3                         & 11.6                         & 6.4                          & 10.0                         & {14.3}                       & 6.9                          & 8.7                          & 11.31                        \\
         Qwen-VL-Max                                  & 15.59                         & 10.7                         & {19.1}                       & 20.0                         & 16.9                         & {12.5}                       & {17.9}                       & 16.4                         & 12.2                         & {21.0}                       & 13.3                         & 14.2                         & 19.8                         & 11.5                         & {20.7}                       & 13.0                         & 17.3                         \\
         Gemini Pro                                   & {17.66}                       & {15.1}                       & 10.7                         & {20.7}                       & {20.1}                       & 11.9                         & 7.5                          & {20.2}                       & {21.1}                       & 16.8                         & 19.1                         & {19.0}                       & {20.0}                       & {14.3}                       & 13.8                         & 17.4                         & {20.8}                       \\
         GPT-4Turbo                                   & 30.26                         & 37.7                         & 33.3                         & 46.4                         & 25.0                         & 28.6                         & 25.3                         & 15.4                         & 27.8                         & 31.9                         & 30.6                         & 29.0                         & 31.9                         & 28.7                         & 37.9                         & 17.4                         & 23.2                         \\ \rowcolor{backblue!75}
         MathFlow$^{\star}$$_{\text{GPT-4V}}$   & \textbf{32.01}  & \textbf{43.9}  & \textbf{38.7}  & \textbf{48.5}  & \textbf{28.2}  & \textbf{29.0}  & \textbf{25.4}  & \textbf{17.7}  & \textbf{30.6}  & \textbf{32.1}  & \textbf{32.9}  & \textbf{30.7}  & \textbf{35.1}  & \textbf{31.5}  & \textbf{40.0}  & \textbf{20.3}  & \textbf{26.0}  \\
         \cmidrule{1-18} ShareGPT4V-13B               & 11.88                         & 7.5                          & 15.5                         & 16.4                         & 10.7                         & 8.9                          & 9.0                          & 11.5                         & 8.9                          & 7.6                          & 11.6                         & 13.0                         & 17.4                         & 10.3                         & 8.6                          & 8.7                          & 12.5                         \\
         SPHINX-MoE                                   & 14.18                         & 7.8                          & 17.9                         & 14.3                         & 15.6                         & 9.5                          & 11.9                         & 12.5                         & 15.6                         & 12.6                         & 16.2                         & 15.6                         & 17.8                         & 13.5                         & 12.1                         & 8.7                          & 16.1                         \\
         InternLM-VL                       & 14.54                         & 9.3                          & 15.5                         & 12.1                         & 15.3                         & 11.3                         & 10.5                         & 14.4                         & {22.2}                       & {19.3}                       & {19.7}                       & 15.6                         & 15.0                         & 11.9                         & 15.5                         & {26.1}                       & 15.5                         \\ 
         Qwen2-VL-72B                                 & 25.90                          & -                            & -                            & -                            & -                            & -                            & -                            & -                            & -                            & -                            & -                            & -                            & -                            & -                            & -                            & -                            & -                            \\ \rowcolor{backblue!75}
         MathFlow$^{\star}$$_{\text{Qwen2-VL-72B}}$ & \ul{28.14} & \ul{38.2} & \ul{36.5} & \ul{47.5} & \ul{27.1} & \ul{23.8} & \ul{20.6} & \ul{22.2} & \ul{21.5} & \ul{27.6} & \ul{19.5} & \ul{24.0} & \ul{28.3} & \ul{27.3} & \ul{33.0} & \ul{35.0} & \ul{18.1} \\
         \bottomrule
      \end{tabular}
   \end{adjustbox}
   \label{tab:tab5}
\vspace{-0.5em}
\end{table*}

\section{Experiments}
\label{sec:experiment}

\subsection{Experimental Setup
}

\noindent
\textbf{Datasets.} We used both publicly
available datasets and internal data to support training.
For the EI caption task, we began by filtering out images with excessively high
resolutions and those unrelated to mathematical content. Ultimately, we
obtained 650,000 image-text pairs.
For the visual-oriented reasoning task, we first selected 40,000 problems from our custom
question bank that contained detailed solution processes. We then
decomposed each problem's solution into individual steps and incorporated them into
prompts, resulting in a final dataset of 130,000 samples. 
{\textit{Please refer to Appendix~\S\ref{Additional_Details_of_Training_Dataset} for more details.}}

\vspace{0.3em}\noindent
\textbf{Baselines.} We benchmarked
MathFlow against SOTA MLLMs. 
For \textit{open-source} models,
we compared MathFlow against Qwen-VL-Plus~\cite{qwen-vl}, Qwen2-VL~\cite{qwen2-vl}, InfiMM-Math~\cite{infimm}, InternVL-2.5~\cite{internvl2.5} and Qwen-VL-Max~\cite{qwen-vl}.
For \textit{close-source} models, we included Claude-sonnet-3.5~\cite{claude-3-5}, Gemini1.5-Pro~\cite{gemini1.5},SPHINX-MoE~\cite{sphinx},  InternLM-XC2~\cite{internlm} and GPT-4V~\cite{gpt4v}.
For \textit{math-specialized model}, we included SVE-Math~\cite{zhang2025open}, MultiMath~\cite{multimath}, and MathLlava~\cite{mathllava}.

\vspace{0.3em}\noindent
\textbf{Implementation Details.} All experiments were conducted in a zero-shot
setting to demonstrate the generalization capabilities of MLLMs in mathematical
reasoning without relying on few-shot prompting or any additional fine-tuning. By
default, we employ th CoT-based) prompting technique~\cite{cot}, which
encourages MLLMs to perform complete reasoning steps for a fine-grained evaluation. We conduct all experiments on NVIDIA A100 GPUs.
During the training of MathFlow, we consistently used DeepSpeed Zero2. In the multi-task pretraining
stage, we adopted a maximum learning rate of 1e-5. In the supervised fine-tuning stage, we used a maximum learning rate
of 5e-6.  
{\textit{Please refer to Appendix~\S\ref{More_Analysis_of_Experiment_Results} for more details.}}

\subsection{Benchmark Results on FlowVerse} \label{sec:flowverse_results}
Tab.~\ref{tab:flowverse_benchmark} compares several state-of-the-art (SOTA)
MLLMs on FlowVerse benchmark, including both closed-source and open-source models.
Notably, in FlowVerse—where Reasoned Property (RP) and Essential Information (EI)
information are manually annotated—MathFlow does not require additional extraction
of these components. Therefore, we do not include MathFlow in direct comparisons
for FlowVerse. We mainly analyze the performance by the FlowVerse-CoT-E
evaluation and derive the following key observations:

\begin{figure}[t]
    \centering
    \includegraphics[width=1\linewidth]{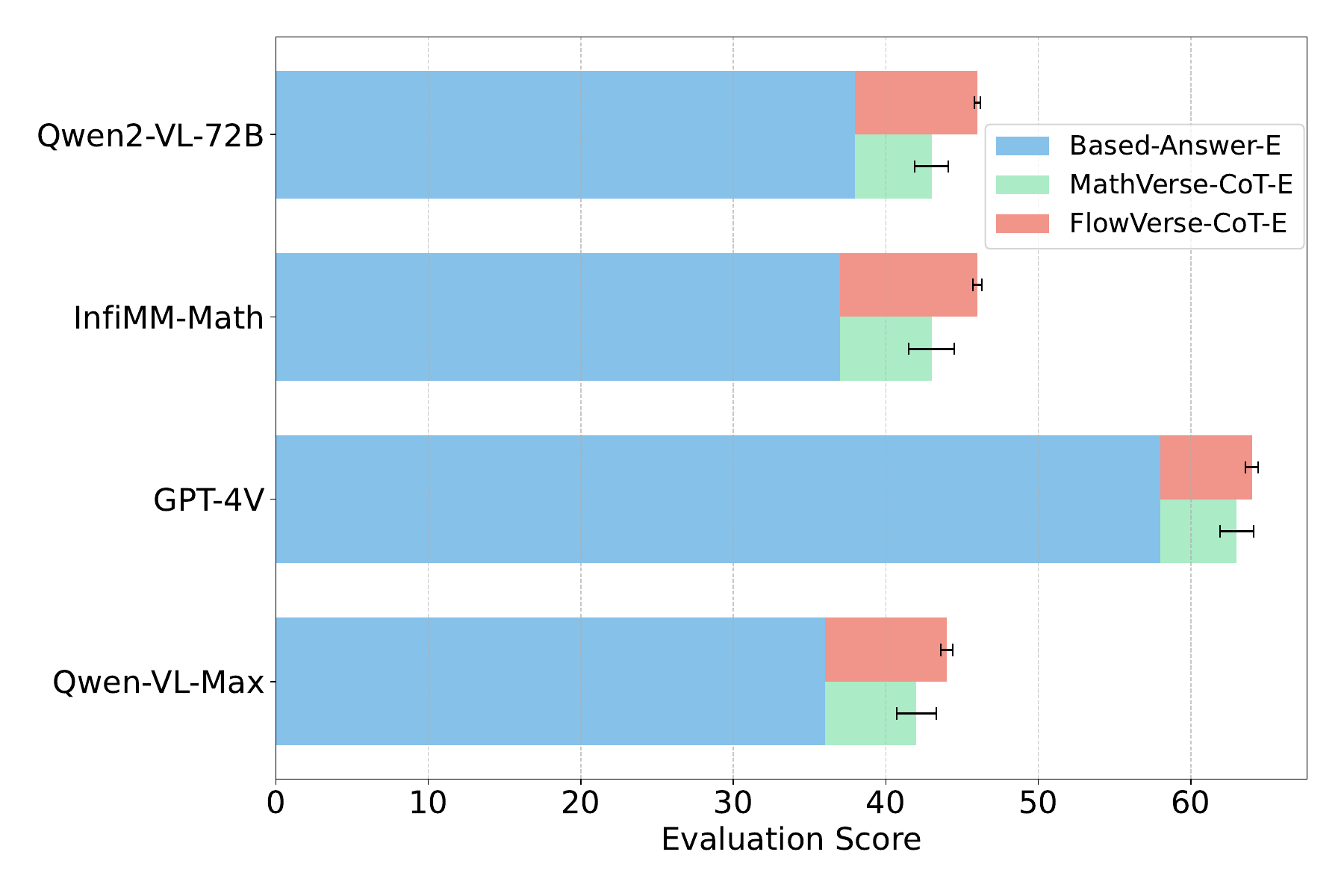}
   \vspace{-2.5em}
    \caption{\textbf{Robustness Comparison of Two Different CoT Evaluation Strategies on FlowVerse$^{\dagger}$.}
    }
    \label{fig:COT_compare}
   \vspace{-0.5em}
\end{figure}

\vspace{0.3em}\noindent
\textbf{MLLMs Rely More on \textit{Reading text} than \textit{Seeing Diagrams}.}
When comparing the Text Centric
and Text Plus versions in FlowVerse, we observe that most MLLMs improve in performance
when visual input is removed, such as a +4.3\% increase for Qwen2-VL-72B.
Conversely, when comparing the Text Centric and
Text Limited versions, we notice a significant drop in performance, such as a -4.1\%
decrease for GPT-4V. These results emphasize the necessity of both comprehensive
textual descriptions and effective visual encoding for robust problem-solving in
MLLMs. {\textit{Please refer to Appendix~\S\ref{More_Detailed_Experiment_Results} for more results.}}

\vspace{0.3em}\noindent
\textbf{MLLMs Significantly Benefit from \textit{EI}.}
Incorporating the EI within diagrams challenges MLLMs to accurately identify and interpret
these critical components visually during mathematical problem-solving. The
Vision Centric results show a significant drop in performance for most MLLMs
compared to their accuracy in the Text Limited version. In other words, we conclude
that a truly robust MLLM must effectively extract EI from diagrams and visual
data. This ability to independently comprehend and interpret visual elements is crucial
for successfully solving visual mathematical problems, indicating that enhanced
perception capabilities are a key factor in improving MLLMs' performance.

\vspace{0.3em}\noindent
\textbf{MLLMs Significantly Benefit from \textit{RP}.} By removing the RP from the question text, we observe a significant decline in accuracy
(Text Limited Version \textit{vs} Vision Dense Version) for most MLLMs, particularly
in open-source models. This indicates that RP contains many inferred
relationships that are essential for guiding the reasoning process and enhancing
the overall problem-solving capabilities of MLLMs. Furthermore, when we convert
RP from the text modality to the image modality, the performance of most MLLMs decreases
substantially. This not only reinforces the observation that MLLMs rely more on
\textit{reading text} than \textit{interpreting diagrams} but also highlights
that MLLMs significantly benefit from the inclusion of \textit{RP}.

\vspace{0.3em}\noindent
\textbf{The FlowVerse-CoT-E Strategy Shows More Robust.} In Fig.~\ref{fig:COT_compare},
we compare two different CoT evaluation strategies: MathVerse-CoT-E and FlowVerse-CoT-E.
Specifically, we evaluated them using both MathVerse-CoT-E and FlowVerse-CoT-E methods
on FlowVerse$^{\dagger}$ dataset, repeating 5 times. The results show that while
MathVerse-CoT-E can demonstrate the effectiveness of fine-grained assessment, its
evaluation lacks robustness due to its reliance on GPT to perform key-step extraction
on MLLM responses, followed by multi-step scoring. In contrast, FlowVerse-CoT-E
uses manually deconstructed steps, resulting in more accurate and robust evaluations.

\begin{table}[t]
    \centering
    \caption{ \textbf{Performance Comparison of MLLMs on MathVerse and FlowVerse$^{\dagger}$
    Datasets.} \textbf{FlowVerse}$^{\dagger}$ indicates the raw version of the
    dataset. ``$MathFlow^{*}$'' denotes the MathFlow pipeline, whose perception model is MathFlow-P-7B. \textit{The full table is provided in Appendix~\S\ref{More_Detailed_Experiment_Results}.}}
    \label{tab:tab3}
   \vspace{-1em}
        \small
        \setlength{\tabcolsep}{7.5pt}
        \begin{tabular}{l|c c}
            \toprule \textbf{Model} & \textbf{MathVerse}           & \textbf{FlowVerse}$^{\dagger}$ \\
            \midrule 
            {Qwen2-VL-72B}                 & 38.9                                    & 52.3                          \\
            {InternVL-2.5-78B}             & 43.2                                    & 54.7                          \\
            {GPT-4V}                       & 54.4                                    & 56.2                         \\
            {Claude-sonnet-3.5}            & 57.4                                    & 64.0                          \\
            {Gemini 2.5-pro}               & 59.9                                    & 68.9                         \\             
            \midrule
            MathFlow$^{\star}$$_{\text{Qwen2-VL-72B}}$              & {48.1} & {58.3}   \\
            MathFlow$^{\star}$$_{\text{GPT-4V}}$     & {56.7}  & {59.3}    \\
            MathFlow$^{\star}$$_{\text{InternVL-2.5-78B}}$         & \ul{56.8} & \ul{60.1}   \\
            MathFlow$^{\star}$$_{\text{Claude-sonnet-3.5}}$     & {60.8} & {68.1}   \\
            MathFlow$^{\star}$$_{\text{Gemini 2.5-pro}}$      & \textbf{62.4}  & \textbf{70.4}    \\

            \bottomrule
        \end{tabular}
    
   \vspace{-0.5em}
\end{table}

\begin{table}[t]
\centering
\caption{\textbf{Structured caption quality evaluation.} MathFlow-P-7B achieves superior performance in extracting both visual elements (EI) and their relationships (RP) from diagrams.}
\vspace{-1em}
\label{tab:captionquality}
\small
\setlength{\tabcolsep}{17pt}
\begin{tabular}{l|cc}
\toprule
\textbf{Model} & \textbf{EI (F1)} & \textbf{RP (F1)} \\
\midrule
GPT-4o-mini     & 84.4\% & 65.3\% \\
GPT-4o          & 87.7\% & 70.1\% \\
Claude-sonnet   & 89.1\% & 72.8\% \\ \rowcolor{backblue!75}
{MathFlow-P-7B} & \textbf{97.2\%} & \textbf{85.6\%} \\
\bottomrule
\end{tabular}
\vspace{-1em}
\end{table}

\begin{figure}[t]
    \centering
    \includegraphics[width=1\linewidth]{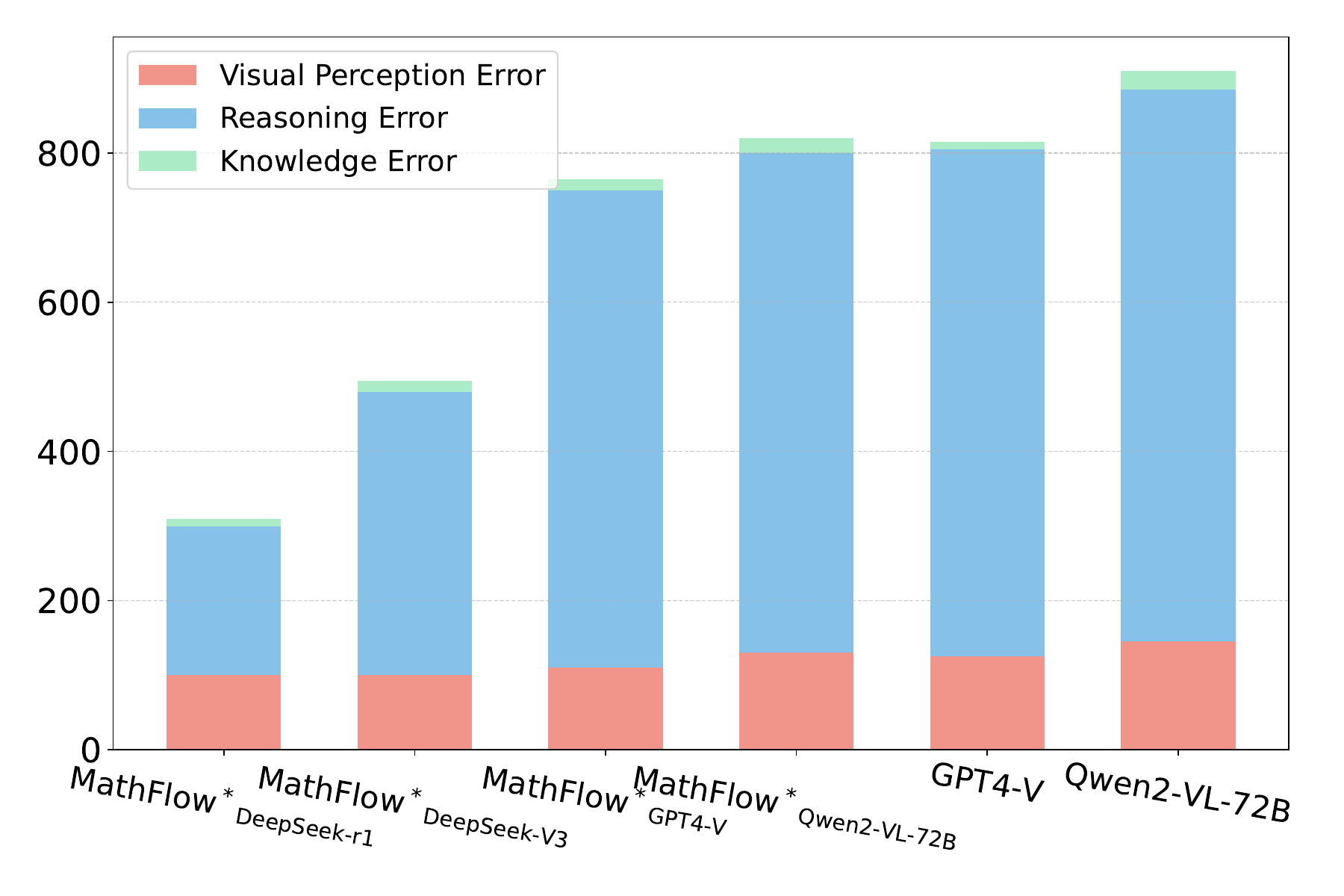}
   \vspace{-2.5em}
    \caption{\textbf{Comparison of Error Distributions Across Models on
    FlowVerse$^{\dagger}$.}}
    \label{fig:type_error}
   \vspace{-1em}
\end{figure}

\subsection{MathFlow Analysis}
\noindent
\textbf{Overall Performance.} 
Tabs.~\ref{tab:flowverse_benchmark}-\ref{tab:tab5} 
show the performance of MLLMs on FlowVerse and MathVerse's testmini set.
MathFlow achieves the highest overall accuracy 
 outperforming other
models like GPT-4V. In terms of CoT-based evaluation (CoT-E), MathFlow$^{\star}$$_{\text{GPT-5}}$
also demonstrates consistent superiority. On the other hand, Tab.~\ref{tab:tab3}
shows the performance comparison of MLLMs on the FlowVerse$^{\dagger}$ and
MathVerse datasets,
where FlowVerse$^{\dagger}$ refers to the raw, unmodified version of FlowVerse.
Notably, MathFlow$^{\star}$$_{\text{GPT-5}}$
achieves the highest accuracy across
both datasets. Furthermore, MathFlow$^{\star}$$_{\text{GPT-V}}$ leads among closed-source models in
arithmetic (Ari), logical reasoning (Log), and geometry (Angle), showing its strength
in diverse mathematical domains. 
These results not only highlight MathFlow's adaptability but also underscore the critical role of perception capabilities.
Furthermore, we also evaluate MathFlow$^{\star}$$_{\text{Qwen2.5-VL-7B}}$ against a set of math-specialized MLLMs on Tab.~\ref{tab:flowverse_benchmark}, MathFlow-P-7B consistently outperforms these baselines across all six problem versions, demonstrating its superior adaptability to diverse multimodal inputs. We attribute these gains to its explicit decoupling of the perception and inference stages, which particularly ensures accurate perception.

\vspace{0.3em}\noindent
\textbf{Error Analysis.} 
Fig.~\ref{fig:type_error} categorizes
error types into three types.
Experts annotated
the responses of multiple models on FlowVerse$^{\dagger}$ to identify these
error types. 
We observe that MathFlow$^{\star}$ not only shows an overall reduction in error rate
compared to the corresponding base model but also exhibits a decrease in visual perception
errors.  Moreover, even with MathFlow$^{\star}_\text{DeepSeek-r1}$, a significant number of perception errors still occur, underscoring our assertion that perception capability is crucial.

\vspace{0.3em}\noindent
\textbf{Structured Caption Evaluation}
We evaluate the quality of visual captioning by assessing how accurately models extract essential elements and their relationships from mathematical diagrams. We adopt the \textit{Structured F1} metric, which quantifies alignment between predicted and ground-truth structured outputs. For each diagram, we decompose the caption into two sets: \textbf{Essential Information (EI)}, such as geometric primitives (\eg, points, lines, circles, and labeled symbols), and \textbf{Reasoned Properties (RP)}, which represent higher-order relationships between elements 
Each predicted EI or RP tuple is matched against human-annotated ground truth using exact set match. The precision, recall, and F1 are computed as:
\begin{equation}
    \text{Precision} = \frac{|\text{Prediction} \cap \text{GroundTruth}|}{|\text{Prediction}|},
\end{equation}
\begin{equation}
    \text{Recall} = \frac{|\text{Prediction} \cap \text{GroundTruth}|}{|\text{GroundTruth}|},
\end{equation}
\begin{equation}
    \text{F1} = \frac{2 \cdot \text{Precision} \cdot \text{Recall}}{\text{Precision} + \text{Recall}}.
\end{equation}

This evaluation is conducted on a curated subset of FlowVerse where high-quality human annotations are available. As shown in Tab.~\ref{tab:captionquality}, MathFlow-P-7B significantly outperforms general-purpose baselines in both aspects. It achieves a Structured F1 of \textbf{97.2\%} on EI extraction, demonstrating highly accurate detection of core visual entities, and \textbf{85.6\%} on RP prediction, indicating strong ability to capture inter-element structural reasoning.

\begin{table}[t]
\footnotesize
\centering
    \caption{\textbf{Ablation Analysis of MathFlow on FlowVerse$^{\dagger}$.} \textbf{Bold}
    numbers indicate the best performance. \textit{{The full table is provided in Appendix~\S\ref{More_Detailed_Experiment_Results}.}}}
    \label{tab:ablation}
   \vspace{-1em}
        \setlength{\tabcolsep}{0.5pt}
        \begin{tabular}{c c c|c}
            \toprule \textbf{\makecell[c]{Perception Model\\for EI}} & \textbf{\makecell[c]{Perception Model\\for RP}} & \textbf{\makecell[c]{Inference\\ Model}} & \textbf{\makecell[c]{COT-E\\(\%)}} \\
            \midrule Qwen2-VL-2B                                     & GPT-4V                                          & GPT-4V                                   & 49.0                               \\
            Qwen2-VL-7B                                              & GPT-4V                                          & GPT-4V                                   & 53.3                               \\
            InternVL2.5-8B                                           & GPT-4V                                          & GPT-4V                                   & 53.8                               \\
            Qwen2.5-VL-7B                                            & GPT-4V                                          & GPT-4V                                   & 53.9                               \\
            MathFlow-P-7B                                            & GPT-4V                                          & GPT-4V                                   & \textbf{58.9}                      \\
            \midrule GPT-4V                                          & Qwen2-VL-2B                                     & GPT-4V                                   & 49.8                               \\
            GPT-4V                                                   & Qwen2-VL-7B                                     & GPT-4V                                   & 52.9                               \\
            GPT-4V                                                   & MathFlow-P-7B                                   & GPT-4V                                   & \textbf{57.3}                      \\
            \midrule GPT-4V                                          & GPT-4V                                          & GPT-4V                                   & 56.2                               \\
            MathFlow-P-7B                                            & MathFlow-P-7B                                   & GPT-4V                                   & \textbf{59.3}                      \\
            MathFlow-P-7B                                            & MathFlow-P-7B                                   & GPT-4o                                   & {69.6}                             \\
            MathFlow-P-7B                                            & MathFlow-P-7B                                   & Gemini 2.5-pro                           & {70.4}                             \\
            MathFlow-P-7B                                            & MathFlow-P-7B                                   & DeepSeek-v3                              & {73.4}                      \\
            MathFlow-P-7B                                            & MathFlow-P-7B                                   & DeepSeek-r1                             & \textbf{75.6}                      \\
            \bottomrule
        \end{tabular}
   \vspace{-0.5em}
\end{table}

\subsection{Ablation Studies} 
\noindent
\textbf{Ablation Analysis of  EI and RP.}
Tab.~\ref{tab:ablation} presents the ablation analysis
of MathFlow on the FlowVerse$^{\dagger}$ dataset, examining the contributions of
different perception models for EI and RP, as well as the impact of varying
inference models. The analysis demonstrates that when GPT-4V is used as the inference
model, MathFlow consistently outperforms other configurations
. Furthermore, we found that substituting other state-of-the-art LLMs as
inference models yielded even higher performance, with MathFlow$^{\star}_\text{DeepSeek-r1}$ reaching its peak
performance.

\vspace{0.3em}\noindent
\textbf{Ablation Analysis of  Perception Model.}
We further compare more models trained with MathFlow as the perception model in Tab.~\ref{tab:ablationb7}, and the MathFlow-trained variant mPLUG-Doc$^{\star}$—in which the EI and RP perception modules share a common backbone but differ in inference model.
In all cases, these models underperformed MathFlow-P-7B, underscoring the efficacy and robustness of the MathFlow pipeline.
This remarkable effectiveness of MathFlow stems from its explicit separation of
perception and inference, allowing for independent optimization of the perception
model and resulting in more accurate visual feature extraction. By converting
complex visual information into textual representations that are readily
Consumable by inference models, MathFlow significantly enhances the reasoning capabilities
of various LLMs/MLLMs. 

\begin{table}[t]
\centering
\caption{\textbf{Performance comparison of various models trained with MathFlow as the perception model.}}
\vspace{-1em}
\scriptsize
\setlength{\tabcolsep}{1.5pt}
\begin{tabular}{l|ccccc}
\toprule 
\textbf{Model} & \textbf{mPLUG-Doc} & \textbf{Qwen} & \textbf{InternVL}  & \textbf{mPLUG-Doc$^{\star}$} & \textbf{MathFlow-P-7B} \\
\midrule GPT-4o & 47.8  & 58.7          & 59.1                            &   62.1  & \textbf{{69.6} }         \\
DS-R1        & 59.3     & 67.8          & 68.6                           &   70.2   & \textbf{{75.6} }       \\
\bottomrule
\end{tabular}
\label{tab:ablationb7}
\vspace{-0.5em}
\end{table}

Notably, we emphasize that MathFlow enables state-of-the-art LLMs to \textbf{solve visual mathematical problems without any additional training}, effectively extending their capabilities beyond pure language processing to visual mathematical reasoning. {\textit{Please refer to Appendix~\S\ref{supsec:mathflow} for more analysis.}}

\section{Conclusion}
\label{sec:conclusion}

We introduce FlowVerse benchmark to investigate the bottlenecks of MLLMs in visual mathematical problem-solving, which reveals the limitations in current MLLMs' ability to accurately perceive and process visual information. Motivated by these findings, we propose MathFlow, which decouples the problem-solving pipeline into perception and inference stages. Given the poor perception performance, we developed MathFlow-P-7B using a two-stage training strategy. 

\section*{Limitations}
In FlowVerse, we initially categorized the collected mathematical problems based
on subjects and subfields that reflect varying degrees of multimodal content.
These classification methods enable a comprehensive evaluation of MLLM
capabilities across different dimensions. However, an additional categorization
by difficulty level—similar to the approach used in datasets like MATH~\cite{MATH}
and WeMath~\cite{wemath}, where problems are sorted by complexity—would provide deeper
insights into model performance. This extra layer of differentiation could
significantly enhance the model's assessment, and we plan to explore this in our
future work.

Furthermore, the issues with FlowVerse primarily stem from the fact that they
are mainly available in English and Chinese. This limitation restricts the model's
applicability to a wider range of languages. By expanding the dataset to include
a greater diversity of languages, we could enhance the multilingual capabilities
of MLLMs and create benchmarks that are more inclusive for users who speak languages
other than English and Chinese.

\section*{Acknowledgments}
This work was supported in part by the National Natural Science Foundation of China (62293554, 62402432) and in part by the Postdoctoral Science Preferential Funding of Zhejiang Province, China (ZJ2025005).

\bibliography{custom}

\clearpage

\appendix

\section*{\huge Appendix}
\startcontents[app]
\section*{Table of Contents}
\printcontents[app]{l}{1}[3]{}

\begin{figure*}[t]
   \centering
   \includegraphics[width=1\linewidth]{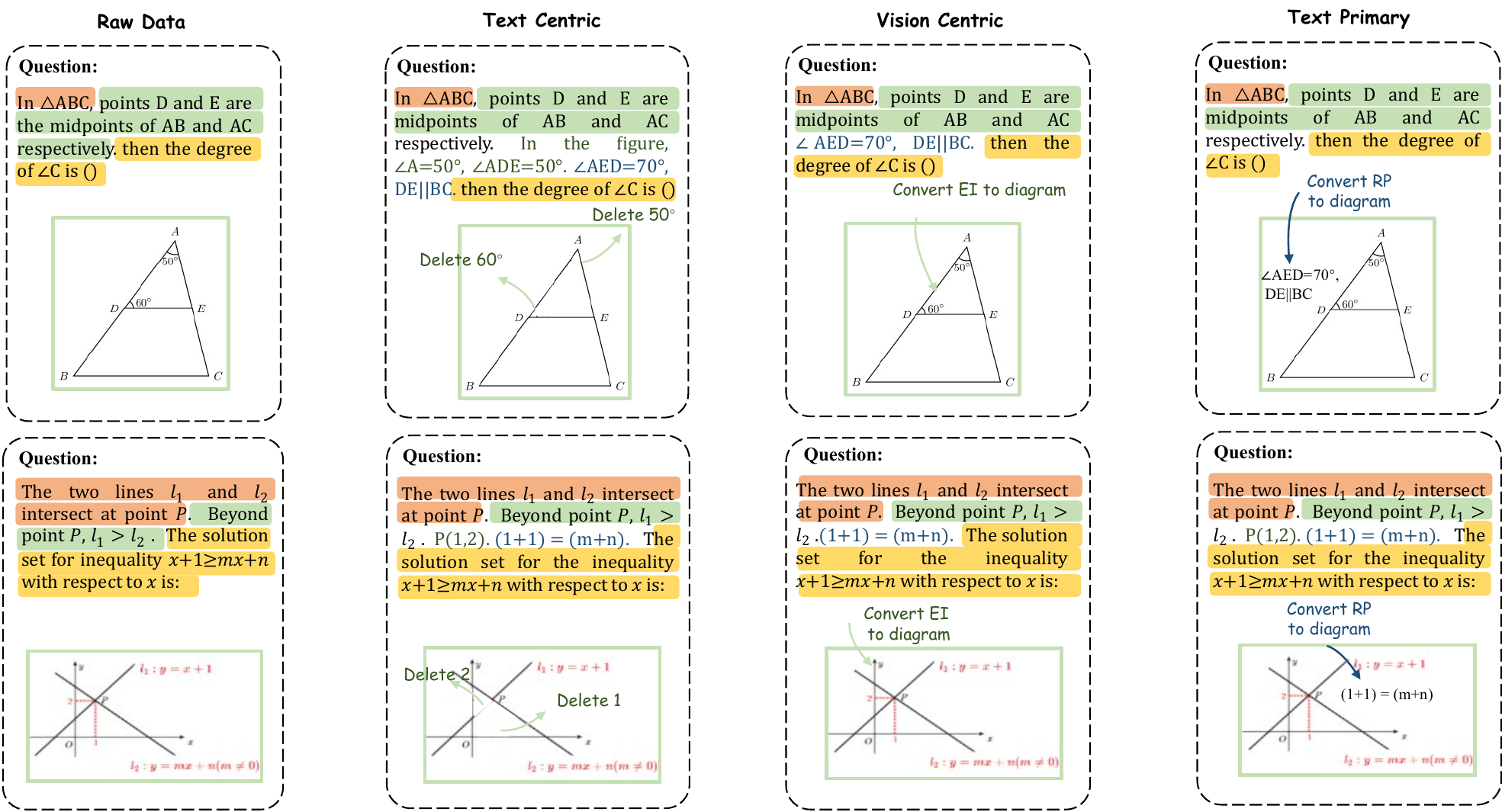}
   \vspace{-2em}
   \caption{\textbf{Manual Modification for EI in FlowVerse.} For the original problems
   shown, we transfer some of the EI from diagrams to question texts (highlighted
   in green) to mark the Vision Centric version.}
   \label{fig:mod_ei}
   \vspace{-0.5em}
\end{figure*}

\begin{figure*}[t]
   \centering
   \includegraphics[width=1\linewidth]{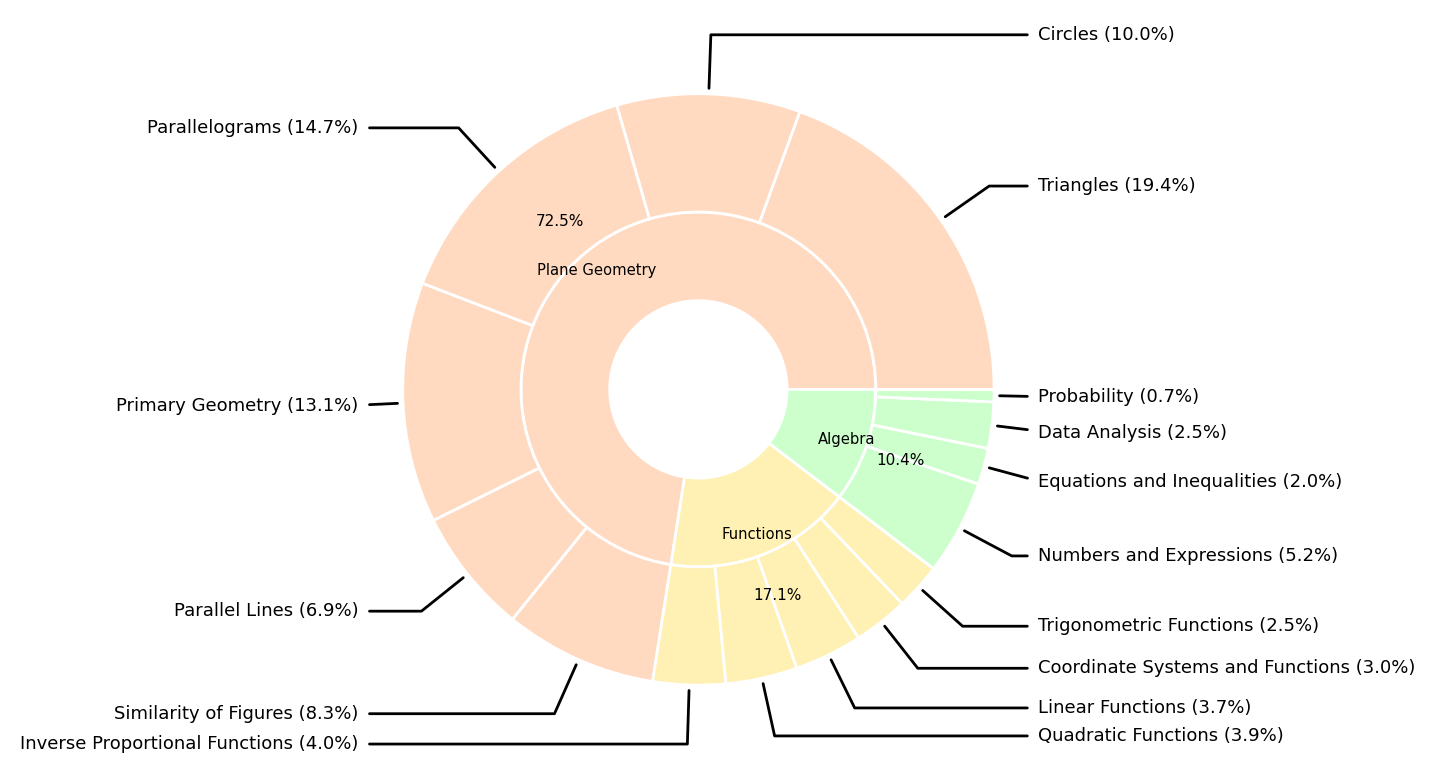}
   \vspace{-2.5em}
   \caption{\textbf{Subject Distribution of FlowVerse.}}
   \label{fig:question_type}
   \vspace{-0.5em}
\end{figure*}

\section{Additional Details of FlowVerse}
\label{supsec:flowverse}
\subsection{Data Curation}
\label{Data_Curation}

\vspace{0.3em}\noindent
\textbf{Data Collection.} We comprehensively collect visual math problems from
internal data$^{1,2}$. Specifically, we focused on collecting problems related to
plane geometry, functions, and algebra. To further enhance the diversity of the
dataset, we introduced additional subcategories within each main category,
capturing more nuanced types of problems. After the initial collection, we
compiled approximately 2,000 visual math problems.

\vspace{0.3em}\noindent
\textbf{Data Categorization and Review.} Initially, human annotators categorized
the collected problems into three main subjects: plane geometry, solid geometry,
and functions. Within each subject, the problems were further subdivided into twelve
fine-grained categories. Subsequently, we meticulously reviewed the dataset,
manually correcting problems with incorrect answers and discarding those that
contained multiple diagrams, visual solutions, or had content overly similar to
other problems. After this careful curation, we preserved 2,000 high-quality
math problems with paired diagrams for FlowVerse, covering a wide range of subjects
and subfields. Notably, to ensure the quality and richness of the dataset, we
excluded problems with minimal Descriptive Information (DI) during the dataset
creation process. Notably, to facilitate evaluation, all proof-based questions have
been reformulated into a question-and-answer format, ensuring consistency and
comparability across different problem types.

\vspace{0.3em}\noindent
\textbf{Transformation of Problem Versions.} Given the four distinct types of information
embedded within the questions, human annotators meticulously transform each
problem into six different versions, as detailed in Sec.~{3.1} of the main paper.
For this purpose, we utilize NetPad$^{3}$ to annotate diagrams for the Text
Dominant, Vision Centric, and Vision Primary versions.
\begin{table}[t]
   \small
   \centering
   \caption{\textbf{Statistics of FlowVerse.}}
   \vspace{-1em}
   \setlength{\tabcolsep}{15pt}
   \begin{tabular}{lr}
      \toprule \textbf{Statistic}          & \textbf{Number}  \\
      \midrule Total Questions             & 2000             \\
      ~- Subjects/subfields                & 3/15             \\
      ~- Multiple-choice questions         & 848 (62.4\%)     \\
      ~- Free-form questions               & 1,152 (37.6\%)   \\
      ~- \textbf{Questions with solutions} & \bf2,000 (100\%) \\
      \midrule Multiple-choice question    &                  \\
      ~- Proportion of answer A            & 171 (20.2\%)     \\
      ~- Proportion of answer B            & 257 (30.3\%)     \\
      ~- Proportion of answer C            & 198 (23.3\%)     \\
      ~- Proportion of answer D            & 200 (23.6\%)     \\
      ~- Proportion of answer E\&F\ \      & 22 (2.6\%)       \\
      \midrule Number of unique images     & 1,906 (95.3\%)   \\
      Number of unique questions           & 2,000 (100\%)    \\
      Number of unique answers             & 561 (28.1\%)     \\
      \midrule Number of English questions & 400 (20\%)       \\
      Number of Chinese questions          & 1600 (80\%)      \\
      \midrule Maximum question length     & 769              \\
      Maximum answer length                & 351              \\
      Average question length              & 104.1            \\
      Average answer length                & 9.9              \\
      \bottomrule
   \end{tabular}
   \label{supp-t5}
   \vspace{-0.5em}
\end{table}

\begin{table}[t]
   \small
   \centering
   \caption{\textbf{Length of Different Problem Versions in FlowVerse.}
   \vspace{-1em}
   }
   \label{supp-t6}
   \setlength{\tabcolsep}{19pt}
   \begin{tabular}{lr}
      \toprule \textbf{Problem Version} & \textbf{Character} \\
      \midrule Text Centric \& Text Plus \\
      ~- Maximum question length        & 962                \\
      ~- Maximum answer length          & 351                \\
      ~- Average question length        & 153.3              \\
      ~- Average answer length          & 9.9                \\
      \midrule Text Limited              \\
      ~- Maximum question length        & 937                \\
      ~- Maximum answer length          & 351                \\
      ~- Average question length        & 123.9              \\
      ~- Average answer length          & 9.9                \\
      \midrule Vision Dense              \\
      ~- Maximum question length        & 688                \\
      ~- Maximum answer length          & 351                \\
      ~- Average question length        & 74.7               \\
      ~- Average answer length          & 9.9                \\
      \midrule Vision Centric            \\
      ~- Maximum question length        & 571                \\
      ~- Maximum answer length          & 351                \\
      ~- Average question length        & 75.8               \\
      ~- Average answer length          & 9.9                \\
      \bottomrule
   \end{tabular}
   \vspace{-0.5em}
\end{table}

For the Text Centric version, as depicted in Fig.~\ref{fig:mod_ei}, where the raw
data presents both EI and RP fully within the diagrams, we selectively remove
portions of this information from the diagrams and integrate them into the text.
This approach emphasizes the textual representation of the key elements.

In the Vision Centric version, NetPad is employed to translate EI from text into
a visual representation, while the corresponding textual information is removed from
the Text Centric version. This transformation aims to shift the balance of
information representation towards visual content.

For the Vision Primary version, NetPad is further utilized to convert RP from text
into visual form based on the Vision Centric version, ultimately creating a
version in which all pertinent information is conveyed solely through diagrams.

\subsection{Subject and Subfield Definition}
\label{Subject_and_Subfield_Definition}
\noindent
\textbf{Plane Geometry.} This foundational area studies the properties and relationships
of points, lines, and surfaces within a two-dimensional plane. It covers key
concepts such as circles, triangles, parallelograms, and parallel lines,
providing a comprehensive context to evaluate the spatial reasoning and logical
deduction abilities of MLLMs. Furthermore, expert annotators have categorized
the problems into twelve fine-grained categories, as illustrated in Fig.~\ref{fig:question_type},
highlighting various dimensions of visual mathematical skills.
\begin{itemize}
   \item \textbf{Circles.} This subfield involves understanding properties and
      relationships associated with circles, such as radius, diameter, tangents,
      arcs, chords, and inscribed angles. Evaluating MLLMs in this area tests their
      ability to comprehend and reason about the unique attributes and properties
      that define circular geometry.
   \item \textbf{Triangles.} This category examines various properties of
      triangles, including side lengths, angles, congruence, similarity, and the
      Pythagorean theorem. Assessing models of learning and language (MLLMs) on
      triangles helps evaluate their understanding of fundamental geometric
      principles and their ability to solve problems involving different types
      of triangles, such as equilateral, isosceles, and scalene triangles.
   \item \textbf{Parallelograms.} This subfield focuses on understanding the
      properties of parallelograms. It covers the relationships between opposite
      sides and angles, the concept that the diagonals bisect each other, and specific
      types of parallelograms such as rectangles, rhombuses, and squares. It
      also tests the understanding of the characteristics that define and
      differentiate these shapes.
   \item \textbf{Parallel Lines.} This subfield focuses on the study of parallel
      lines, especially the angles created when they are intersected by a transversal.
      Key angles to examine include alternate interior angles and corresponding
      angles. It is essential to evaluate the ability of MLLMs to apply these properties
      in order to solve complex geometric problems.
   \item \textbf{Similarity of Figures.} This area examines the criteria for
      similarity between geometric figures, such as side ratios and corresponding
      angles. It requires MLLMs to apply proportional reasoning and recognize
      conditions under which two figures are similar, testing their understanding
      of geometric relationships in a broader context.
   \item \textbf{Primary Geometry.} This subfield deals with fundamental
      concepts such as points, lines, and basic shapes. It lays the foundation for
      understanding more complex geometrical relationships, testing MLLMs' ability
      to grasp and reason with the basic building blocks of geometry.
\end{itemize}

\vspace{0.3em}\noindent
\textbf{Functions.} This encompasses analyzing mathematical relationships between
variables, ranging from simple evaluations of function values to more complex
tasks like examining different function types and their behaviors. We assess MLLMs'
capabilities using four categories of function problems.
\begin{itemize}
   \item \textbf{Inverse Proportional Functions:} This category involves
      functions where the product of two variables remains constant. Evaluating MLLMs
      on inverse proportional functions helps assess their understanding of non-linear
      relationships and their ability to apply these principles in problem-solving.
   \item \textbf{Quadratic Functions:} Problems in this category include
      analyzing the properties of quadratic equations, such as identifying the vertex,
      roots, axis of symmetry, and the impact of changing coefficients. MLLMs are
      tested on their ability to understand and manipulate quadratic expressions
      to solve for function behavior.
   \item \textbf{Linear Functions:} Linear functions represent relationships
      with a constant rate of change. This category focuses on understanding slope
      and y-intercept and interpreting graphical representations of linear equations,
      which are essential for evaluating MLLMs' grasp of fundamental linear relationships.
   \item \textbf{Coordinate Systems and Functions:} This subfield deals with
      understanding how functions are represented within coordinate planes, including
      plotting points, interpreting graphs, and understanding shifts and transformations
      of basic functions. It assesses MLLMs' skills in translating functional relationships
      into visual formats.
   \item \textbf{Trigonometric Functions:} Problems in this category involve
      understanding trigonometric relationships, such as sine, cosine, and tangent,
      and their applications to solve problems involving angles and periodic
      phenomena. Evaluating MLLMs on these functions helps gauge their understanding
      of trigonometric concepts and their application in different scenarios.
\end{itemize}
\vspace{0.3em}\noindent
\textbf{Algebra.} encompasses a wide range of topics that focus on understanding
mathematical relationships through symbols and expressions. It involves solving problems
related to probability, data analysis, equations, and numerical expressions. We evaluate
MLLMs' skills using four distinct categories within algebra:
\begin{itemize}
   \item \textbf{Probability:} This category covers fundamental probability
      concepts, including calculating the likelihood of events, understanding
      random experiments, and applying basic probability laws. It assesses MLLMs'
      ability to handle uncertainty and predict outcomes based on given conditions.

   \item \textbf{Data Analysis:} Problems in this category involve interpreting
      and analyzing data sets, including calculating measures such as mean,
      median, mode, and range, and understanding data representations like charts
      and graphs. Evaluating MLLMs in this area tests their ability to draw
      meaningful conclusions from quantitative data.

   \item \textbf{Equations and Inequalities:} This subfield includes solving
      linear and non-linear equations, as well as inequalities involving one or more
      variables. MLLMs are evaluated on their ability to manipulate and solve
      algebraic expressions and apply them to real-world scenarios.

   \item \textbf{Numbers and Expressions:} This category deals with
      understanding numerical properties, simplifying algebraic expressions, and
      performing arithmetic operations. It tests MLLMs' grasp of numerical
      relationships and their ability to work with algebraic symbols effectively.
\end{itemize}

\subsection{Detailed Statistics of FlowVerse}
\label{Detailed_Statistics_of_FlowVerse}
\noindent
\textbf{More Data Statistics.} In Tab.~\ref{supp-t5}, we present a detailed breakdown
of the FlowVerse dataset. It is important to note that all problems in FlowVerse
are sourced from internet data rather than other datasets, and each question is annotated
manually. Additionally, our evaluation aims to effectively showcase the reasoning
capabilities of MLLMs with moderate-level mathematical knowledge without restricting
their performance through overly complex domain-specific theorems or extensive commonsense
knowledge. Consequently, we focus on problems at the high school level,
deliberately excluding advanced college-level topics such as calculus and graph
theory.

\begin{figure*}[t]
   \centering
    \includegraphics[width=\linewidth]{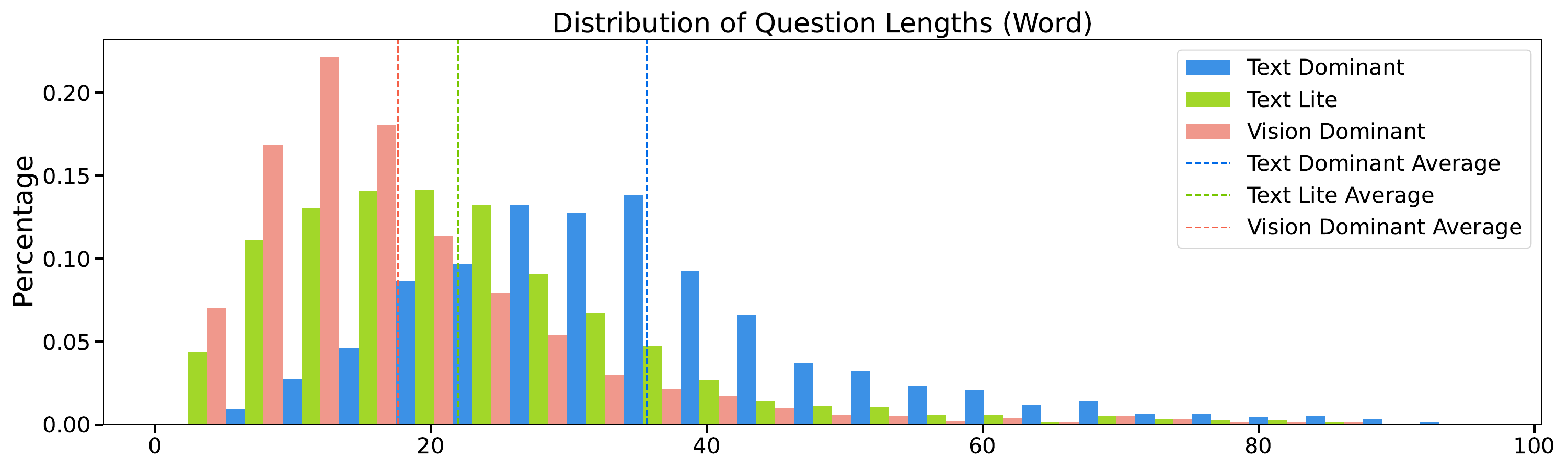}
    \vspace{-2.5em}
   \caption{\textbf{Distribution of Question Length for Four Problem Versions.} We
   present the distribution of question length for the four problem versions,
   with the horizontal axis representing question length in characters and the
   vertical axis depicting the corresponding probability distribution.}
   \label{sup-length_hist}
   \vspace{-0.5em}
\end{figure*}

\vspace{0.3em}\noindent
\textbf{Problem Length Variance.} In Tab.~\ref{supp-t6}, we present the
variations in question and answer lengths across five versions of problems in
FlowVerse, excluding the Vision Primary category, as it contains only the ``Only
Question" component.
By removing pre-defined components such as Descriptive Information (DI), Reasoned
Property (RP), and Essential Information (EI), the maximum and average lengths of
questions decrease progressively, whereas the answer lengths remain unaffected. Fig.~\ref{sup-length_hist}
visualizes the character-level variation in question length for four problem
versions: Text Centric (blue), Text Limited (green), Vision Centric (red), and
Vision Dense (yellow). As DI, EI, and RP are sequentially omitted from the Text Centric
version, we observe a clear downward trend in both the distribution of questions
lengths and their average values.

\subsection{Details of FlowVerse CoT Evaluation}
\label{Details_of_Evaluation_on_FlowVerse}
\noindent
\textbf{Prompt Design for Response Generation.} We employ two distinct types of prompts
for free-form and multiple-choice questions, respectively, as per the guidelines
established by MathVerse~\cite{mathverse}, which are detailed in Tab.~\ref{supp-t1}.
To effectively elicit the Chain-of-Thought (CoT) reasoning capabilities of MLLMs,
we also incorporate the phrase \textit{``first conduct reasoning''} to encourage
a more structured and logical approach to problem-solving.

\vspace{0.3em}\noindent
\textbf{Prompt for CoT Evaluation.} As discussed earlier, our proposed Chain-of-Thought
evaluation, termed FlowVerse-CoT-E, is specifically designed to assess the
reasoning depth and perceptual accuracy of MLLMs in visual mathematics problems.
FlowVerse-CoT-E utilizes a structured prompt that guides the model step-by-step through
the problem-solving process, ensuring a more transparent evaluation of the model's
ability to logically infer and connect various aspects of the problem. This evaluation
mechanism distinguishes between different reasoning paths taken by the model,
thereby providing a nuanced understanding of its intermediate problem-solving
capabilities, as shown in Tab.~\ref{supp-t2}.

\begin{figure}[t]
   \centering
   \includegraphics[width=1\linewidth]{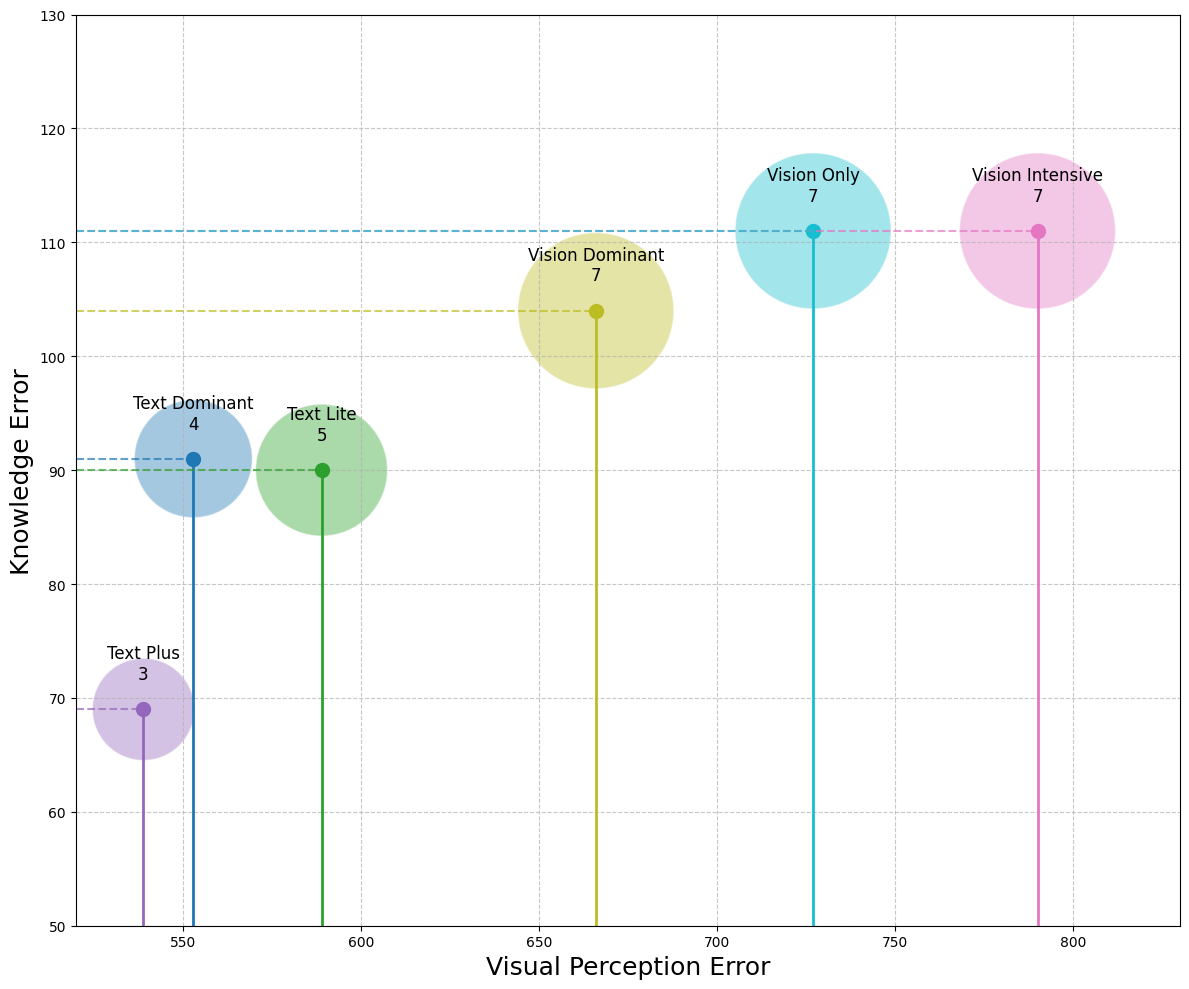}
   \vspace{-2em}
   \caption{\textbf{Visualization of Different Error Type across Different
   Versions using GPT-4 on FlowVerse.} The horizontal axis represents different problem
   versions, while the vertical axis indicates the error types. The radius of each
   bubble corresponds to the number of visual perception errors, with smaller
   radii indicating fewer visual perception errors.}
   \label{fig:bubble}
   \vspace{-0.5em}
\end{figure}

\begin{figure*}[t]
   \centering
   \includegraphics[width=1.0\textwidth]{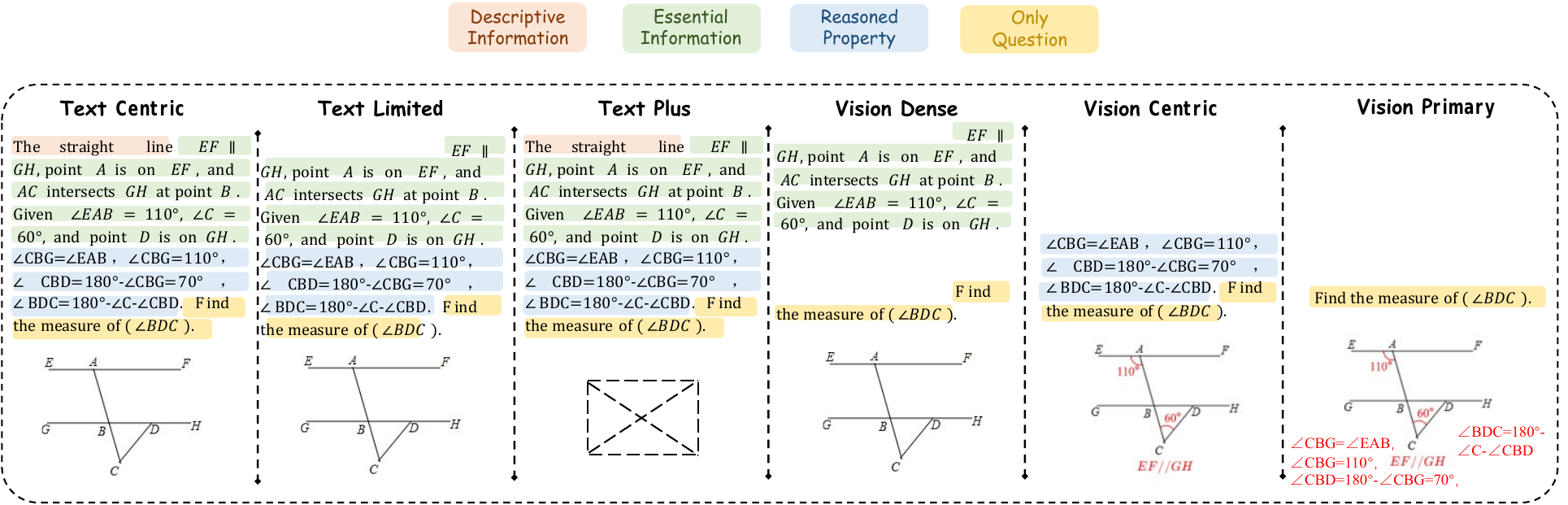}
   \vspace{-2em}
   \caption{\textbf{Comparison of Six Problem Versions in FlowVerse.}}
   \label{case1}
\end{figure*}

\begin{figure*}[t]
   \centering
   \includegraphics[width=1.0\textwidth]{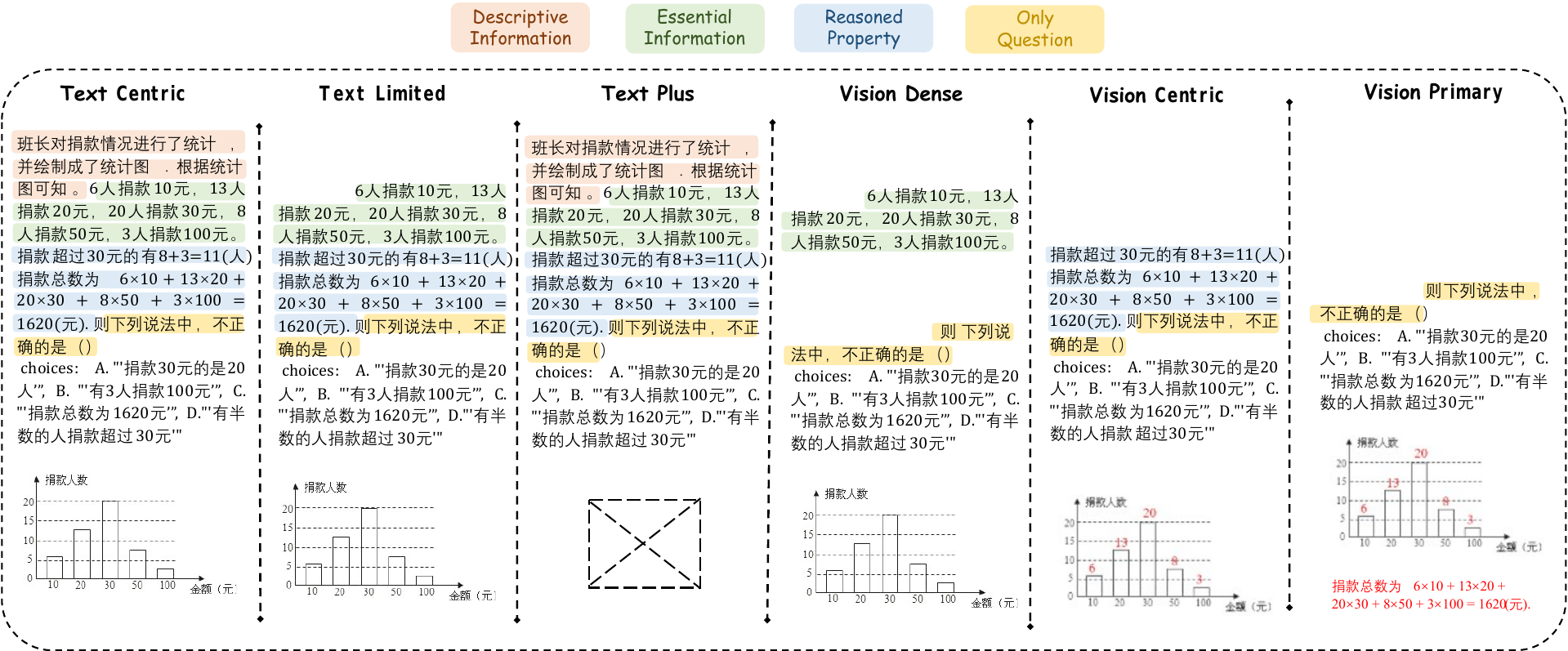}
   \vspace{-2em}
   \caption{\textbf{Comparison of Six Problem Versions in FlowVerse.}}
   \label{case2}
\end{figure*}

\begin{figure*}[t]
   \centering
   \includegraphics[width=1.0\textwidth]{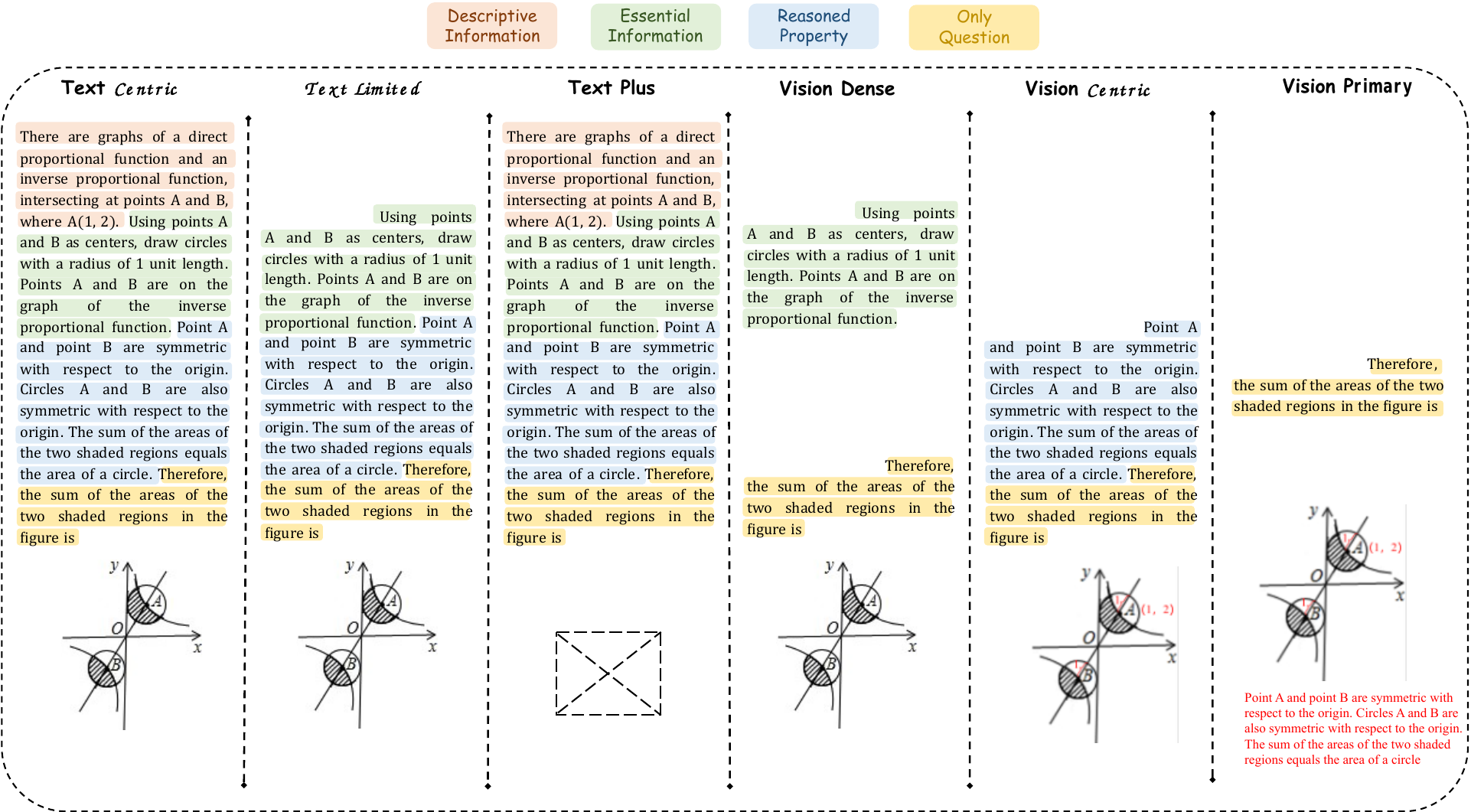}
   \vspace{-2em}
   \caption{\textbf{Comparison of Six Problem Versions in FlowVerse.}}
   \label{case3}
   \vspace{-0.5em}
\end{figure*}

\vspace{0.3em}\noindent
\textbf{Error Analysis of GPT-4 on FlowVerse.} Fig.~\ref{fig:bubble} illustrates
GPT-4V's Knowledge Error, Reasoning Error, and Visual Perception Error across different
versions of problem representations. The Text Plus version demonstrates the
lowest Knowledge and Reasoning Errors, along with a smaller Visual Perception Error,
as indicated by its relatively small bubble radius. This makes it the most effective
approach for minimizing overall errors. In contrast, the Vision Centric, Vision
Only, and Vision Dense versions show significantly higher Reasoning and Knowledge
Errors, accompanied by the largest bubble radii, indicating substantial Visual Perception
Error. This suggests that an over-reliance on visual information increases the
complexity of both perception and reasoning. The Text Centric and Text Limited versions
fall in between, exhibiting moderate error rates and bubble sizes, which implies
a slight increase in perception difficulty as the representation shifts away from
textual information. Overall, the analysis highlights that enriched textual representation,
such as in the Text Plus version, is crucial for achieving optimal performance in
minimizing Knowledge, Reasoning, and Visual Perception errors, underscoring the
importance of a balanced information representation for effective problem-solving.
In conclusion, the result further confirms that the limited capabilities of
prior methods to extract information during the perception stage restrict the
overall problem-solving performance, which also guides us in the development of the
modular problem-solving pipeline, MathFlow.

\begin{table*}
   [!t]
   \centering
   \small
   \caption{\textbf{Input Prompt of MLLMs for Response Generation.}}
   \vspace{-1em}
   \begin{tabular}{p{0.36\textwidth}p{0.52\textwidth}}
      \toprule \textbf{Question}                         & \textbf{Prompt}                                                                                                                   \\
      \midrule \multirow{4}{*}{Free-form Question}       & Please first conduct reasoning, and then answer the question and provide the final value, \eg, 1, 2.5, 300, at the end.          \\
                                                         & -- \textbf{Question: $\{question\}$}                                                                                              \\
      \midrule \multirow{4}{*}{Multiple-choice Question} & Please first conduct reasoning, and then answer the question and provide the correct option letter, \eg, A, B, C, D, at the end. \\
                                                         & -- \textbf{Question: $\{question\}$}                                                                                              \\
      \bottomrule
   \end{tabular}
   \label{supp-t1}
   \vspace{-1em}
\end{table*}

\begin{table*}
   [!t]
   \centering
   \small
   \caption{\textbf{Configuration for the FlowVerse-CoT-E.}}
   \vspace{-1em}
   \begin{tabular}{p{0.36\textwidth}p{0.52\textwidth}}
      \toprule \textbf{Input}                                      & \textbf{Prompt}                                                                                                                                                                          \\
      \midrule \multirow{10}{*}{Question \& Sequence of Solutions} & You will be provided with a visual mathematics problem along with a detailed solution procedure. Your task is to solve the problem by utilizing the provided solution steps as guidance. \\
                                                                   & Here are examples:                                                                                                                                                                       \\
                                                                   & -- Question: XXX                                                                                                                                                                         \\
                                                                   & -- Solution: 1. XXX 2. XXX 3. XXX                                                                                                                                                        \\
                                                                   & -- Final Solution: XXX

Here is what you need to solve:                                                                                                                                  \\
                                                                   & -- \textbf{Question: $\{question\}$}                                                                                                                                                     \\
                                                                   & -- \textbf{Solution: $\{solution\}$}                                                                                                                                                     \\
                                                                   & -- \textbf{Final Solution:}                                                                                                                                                              \\
      \bottomrule
   \end{tabular}
   \label{supp-t2}
   \vspace{-0.5em}
\end{table*}

\subsection{Qualitative Examples}
\label{Qualitative_Examples}

Figs.~\ref{case1}-\ref{case3} illustrate the differences across six versions of problem
representations: Text Centric, Text Limited, Text Plus, Vision Centric, Vision
Intensive, and Vision Primary. Each version provides a distinct balance between
textual and visual information, thereby influencing the accessibility and interpretability
of the problem.

The Text Centric version includes detailed descriptive information along with key
reasoning steps in the text, while the Text Limited version reduces some
descriptive elements to focus on core content. The Text Plus version enriches this
further with added visual context for better comprehension. On the other hand, the
Vision Centric and Vision Dense versions emphasize visual elements by gradually minimizing
textual information, whereas the Vision Primary version conveys almost all the
information through diagrams with minimal textual support.

These examples highlight the impact of information representation on the model's
ability to comprehend and solve problems, demonstrating how variations in text and
visual content can influence both perception and reasoning capabilities.

\section{Additional Details of Training Dataset }
\label{Additional_Details_of_Training_Dataset}

\subsection{Details of MathFlow-RP}
\label{Details_of_MathFlow-RP}
As mentioned in Sec.\textcolor{black}{3.2}, We designed MathFlow-RP to enhance
the reasoned property of MathFlow during the multi-task pretraining stage for the
RP caption task, which focuses on extracting higher-level abstractions and
relationships, requiring the model to engage in abstract reasoning—such as deducing
relationships between geometric shapes or identifying intersection points in a function.

\vspace{0.3em}\noindent
\textbf{Data Collection.} As previously described, the MathFlow-RP dataset is meticulously
curated to enhance the model’s response quality and adapt it more effectively to
the current task context. This dataset consists of 40,000 questions, each accompanied
by a detailed solution, thus providing a robust foundation for enhancing the
model's capability to comprehend complex visual mathematical reasoning.
\begin{figure}[t]
    \centering
    \includegraphics[width=1\linewidth]{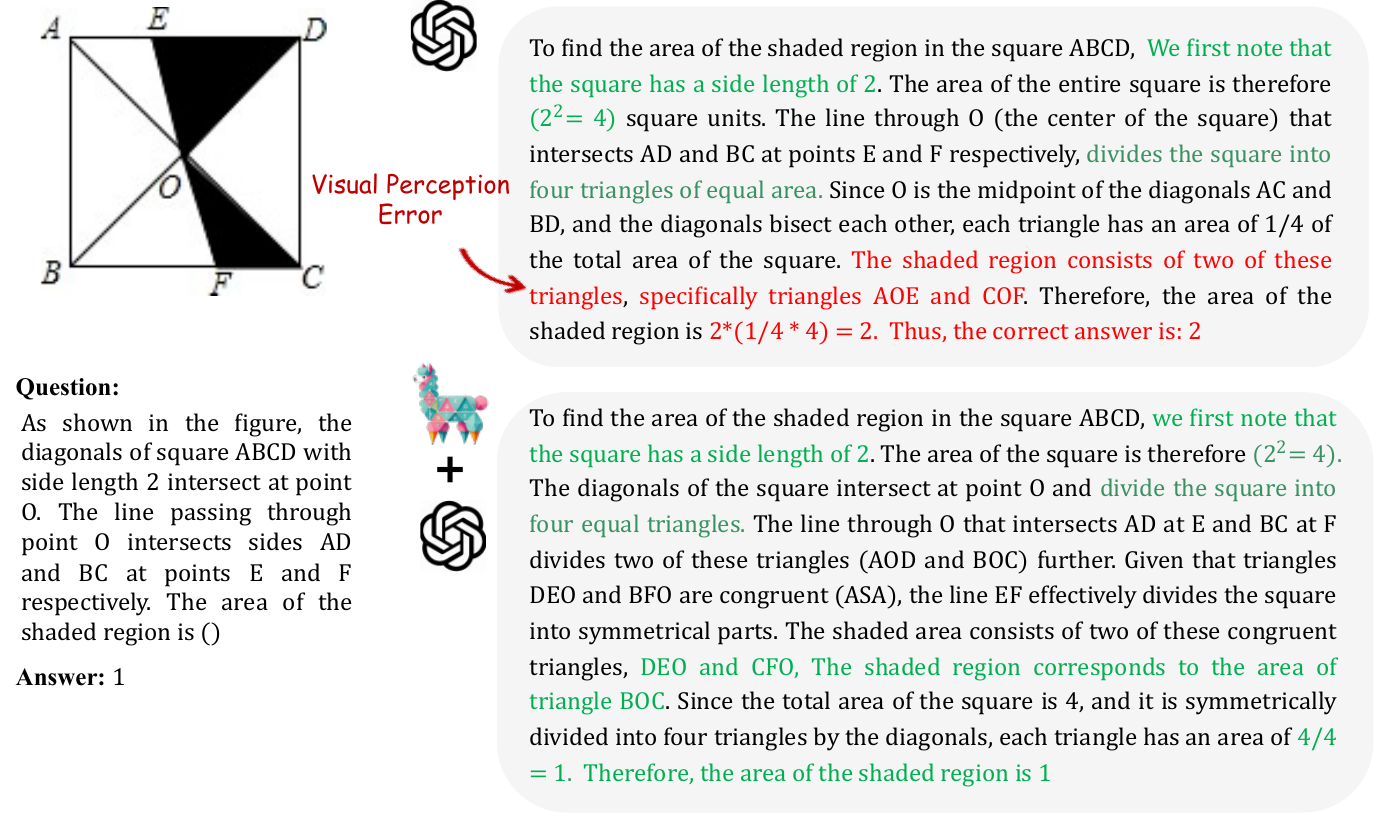}
    \vspace{-2em}
    \caption{\textbf{Problem-solving Comparison of MathFlow$^{\star}$$_{\text{\textbf{GPT-4V}}}$
    and GPT-4V.} 
 The results indicate that MathFlow-P-7B demonstrates significantly better
capability in understanding and addressing complex mathematical problems,
showcasing more effective perception and inference abilities compared to GPT-4V.}
    \label{fig:problem_case}
   \vspace{-0.5em}
\end{figure}

\vspace{0.3em}\noindent
\textbf{Data Annotation and Filtering.} Specifically, as illustrated in Fig.~\ref{p-RP},
Qwen2.5-72B~\cite{qwen2.5} is employed to extract the key steps from these solutions,
given the prohibitive cost of utilizing GPT-4 for the extraction across such a
large dataset. As a result, we were able to compile a total of 13,000 annotated
instances, which serve as the core training data for enhancing the model’s problem-solving
capabilities. Notably, since some problems involve multiple questions, we divide
them into different instances for clarity.

\vspace{0.3em}\noindent
\textbf{Training Strategy.} Building on this approach, the extracted solution
steps are sequentially appended to the original question, enabling the model to predict
the subsequent solution step based on the cumulative information. This iterative
process facilitates a chain-of-thought reasoning mechanism by progressively
integrating previous steps into the prompt, allowing the model to incrementally build
upon earlier conclusions. Specifically, each step in the solution is derived using
pertinent properties, such as congruent triangles or properties of parallel
lines, ensuring that every next prediction is grounded in logical progression.
By extracting, labeling, and integrating key steps, the model is trained to emulate
human-like deductive reasoning, ultimately enhancing its problem-solving
accuracy for complex visual mathematical tasks.

\begin{figure*}[ht]
   \centering
   \includegraphics[width=1.0\textwidth]{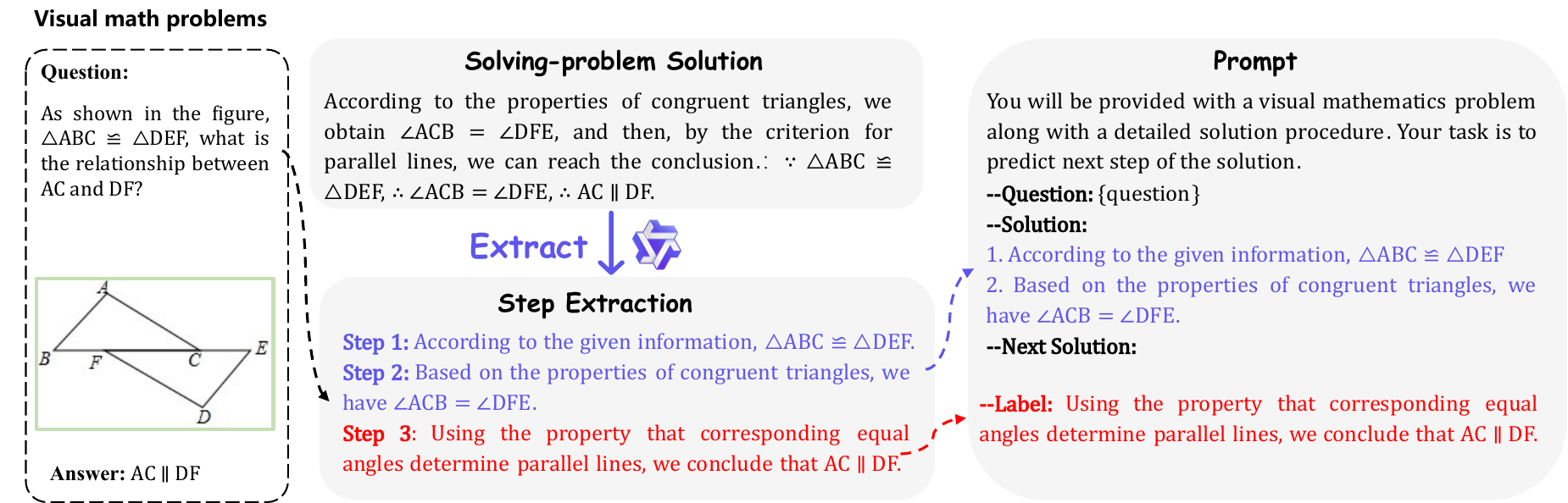}
   \vspace{-2em}
   \caption{\textbf{Data Annotation of the MathFlow-RP} We first employ Qwen2.5-72B
   to extract the corresponding steps from the solving-problem solution, then select
   step \textit{N} as the target for prediction. Subsequently, the preceding
   \textit{N-1} steps are provided as input within the prompt, enabling the MLLM
   to predict the next step based on this sequential context.}
   \label{p-RP}
\end{figure*}

\subsection{Details of MathFlow-SFT}
\label{Details_of_MathFlow-SFT}

As mentioned in Sec.~\ref{sec:mathflow}, the MathFlow-SFT (Supervised Fine-Tuning) dataset is
meticulously curated to enhance the model’s response quality and adapt it more
effectively to the current task context during the supervised fine-tuning stage
of MathFlow. The dataset focuses on extracting and refining the key visual and textual
elements necessary for accurate mathematical reasoning, with specific emphasis
on retaining only the essential information to clearly define solution steps.

\vspace{0.3em}\noindent
\textbf{Data Collection.} The MathFlow-SFT dataset samples visual mathematical problems
with detailed solutions, which are sourced from diverse educational resources, including
textbooks, standardized exam questions, and synthetic problems specifically designed
for visual challenges. Also, MathFlow-SFT is collected to ensure coverage across
different types of mathematical reasoning, such as plane geometry, algebra, and
trigonometry, to provide a comprehensive training base.

\vspace{0.3em}\noindent
\textbf{Annotation of Key Elements.} To ensure that the model effectively
perceives and interprets diagrams, as Fig.~\ref{p-sft} shows, we manually annotated
the critical components of each problem based on both the descriptive information
provided in the problem statement and the corresponding solution. The
annotations include:

\begin{itemize}
   \item Geometric Elements: Labels for points, lines, angles, and other geometric
      properties.

   \item Relationships: Specific relationships, such as congruency or parallelism,
      are highlighted to help the model understand the underlying structure.

   \item Reasoned Property Extraction: Abstract properties inferred from the
      given diagrams and textual context are labeled, emphasizing the reasoning process
      needed to progress through each solution.
\end{itemize}

\begin{figure*}[ht]
   \centering
   \includegraphics[width=1.0\textwidth]{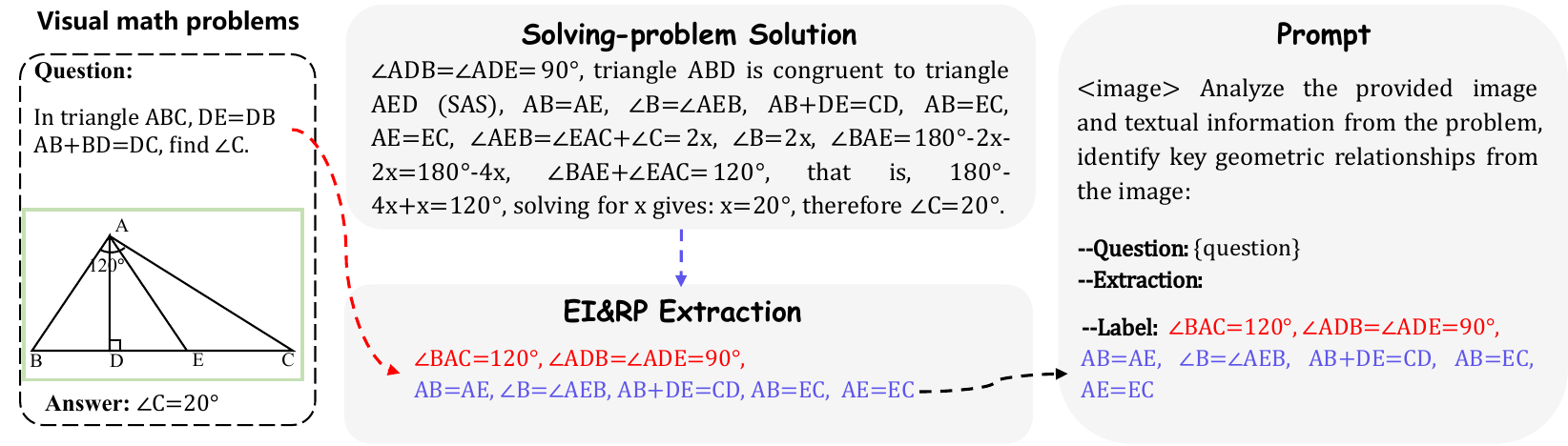}
   \vspace{-2em}
   \caption{\textbf{Data Annotation of the MathFlow-SFT.}We manually extract the
   corresponding EI and RP from the solving-problem solution and associated
   diagram. In this representation, the red-highlighted portions indicate EI, while
   the blue-highlighted sections represent RP.}
   \label{p-sft}
\end{figure*}

\begin{table}[t!]
   \centering
   \caption{\textbf{Generate Config of MathFlow.}}
    \small
   \vspace{-1em}
   \setlength{\tabcolsep}{24pt}
   \begin{tabular}{lr}
      \toprule \textbf{Hyper Parameter} & \textbf{Value} \\
      \midrule \textit{Temperature}     & 0.3            \\
      \textit{TopP}                     & 0.7            \\
      \textit{TopK}                     & 1.0            \\
      \textit{Repetition Penalty}       & 1.0            \\
      \textit{Num of Beams}             & 1.0            \\
      \bottomrule
   \end{tabular}
   \label{sup_Generate_Config}
   \vspace{-0.5em}
\end{table}

\begin{table*}[!t]
\small
\centering
\caption{\textbf{Mathematical Evaluation on Six Problem Versions in FlowVerse}. 
DI, EI, RP, OQ refer to the {textual} or {visual} Descriptive Information, Reasoned Property, Essential Information, Only Question, respectively. 
The Text Plus Version does not involve image input. 
``CoT-E'' or ``Acc'' denotes whether to employ the FlowVerse-CoT-E strategy or not. 
The highest accuracy for each group of MLLMs is marked in \textbf{bold}. 
}
\vspace{-1em}
\begin{adjustbox}{width=\linewidth}
\begin{tabular}{l|C{0.9cm}C{0.9cm}|C{0.9cm}C{0.9cm}|C{0.9cm}C{0.9cm}|C{0.9cm}C{0.9cm}|C{0.9cm}C{0.9cm}|C{0.9cm}C{0.9cm}|C{0.9cm}C{0.9cm}}
\toprule
\multirow{4}*{\makecell*[l]{\large \textbf{Model}}}    
& \multicolumn{2}{c|}{\textbf{All}} 
& \multicolumn{2}{c|}{\textbf{Text Centric}} 
& \multicolumn{2}{c|}{\textbf{Text Limited}}
& \multicolumn{2}{c|}{\textbf{Text Plus}}
& \multicolumn{2}{c|}{\textbf{Vision Dense}}
& \multicolumn{2}{c|}{\textbf{Vision Centric}}
& \multicolumn{2}{c}{\textbf{Vision Primary}} \\
~ & \multicolumn{2}{c|}{~} 
& \multicolumn{2}{c|}{DI+RP+EI+OQ} 
& \multicolumn{2}{c|}{RP+EI+OQ} 
& \multicolumn{2}{c|}{DI+RP+EI+OQ} 
& \multicolumn{2}{c|}{EI+OQ} 
& \multicolumn{2}{c|}{RP+EI+OQ} 
& \multicolumn{2}{c}{RP+EI+OQ} \\
\cmidrule{2-15}
~ & CoT-E & Acc & CoT-E & Acc & CoT-E & Acc & CoT-E & Acc & CoT-E & Acc & CoT-E & Acc & CoT-E & Acc \\
\midrule
\multicolumn{15}{c}{\textit{Open-source MLLMs}}\\
\midrule
InfiMM-Math-7B &36.5&28.8&43.8&38.1&40.6&36.7&46.1&40.1&28.8&15.4&39.6&30.3&26.1&23.2\\
InternVL2.5-8B &44.7&41.0&49.2&41.3&40.5&38.4&49.6&42.7&38.4&20.2&41.0&35.9&35.8&33.9\\
Qwen2.5-VL-7B  &53.8&42.2&60.1&52.8&58.9&51.3&62.0&55.0&45.0&31.0&50.8&46.3&48.1&45.3\\
VLM-R1-7B$^{\dagger}$ &50.7&41.2&59.0&54.2&57.9&49.8&65.5&58.9&36.2&24.5&46.1&37.8&30.6&26.1\\
Qwen2-VL-72B   &52.3&48.6&59.4&47.3&54.3&45.7&63.7&50.0&40.8&25.3&50.9&42.1&47.6&37.0\\
InternVL2.5-78B&54.7&50.1&66.1&62.7&64.1&60.3&67.8&64.7&48.7&34.3&63.0&58.8&59.6&57.7\\ \rowcolor{backblue!75}
MathFlow$^{\star}_{\text{Qwen2.5\!-\!VL\!-\!7B}}$
&57.0&46.0&62.0&53.8&60.8&52.2&64.2&56.0&49.0&39.1&54.5&51.6&52.0&51.5\\
\midrule
\multicolumn{15}{c}{\textit{Math-specialized MLLMs}}\\
\midrule
Math\!-\!LLaVA\!-\!13B &38.0&29.9&45.1&39.3&44.4&37.4&--&--&36.2&18.6&41.7&35.9&37.0&34.2\\
MultiMath-7B &44.0&34.2&50.6&44.8&49.9&42.9&--&--&41.7&22.1&47.2&40.4&39.7&38.8\\
SVE-Math-Qwen2.5-7B &46.0&39.4&53.1&47.3&53.4&45.8&--&--&44.2&28.6&48.9&44.2&45.8&42.0\\
\midrule
\multicolumn{15}{c}{\textit{Closed-source MLLMs}}\\
\midrule
Qwen-VL-Plus  
& 34.4 & 27.1 
& 38.1 & 24.6 
& 35.7 & 28.6 
& 41.7 & 35.6 
& 28.5 & 20.9 
& 32.6 & 26.0 
& 30.0 & 22.2 \\
Gemini-Pro
& 39.0 & 23.5 
& 44.3 & 36.5 
& 40.2 & 33.8 
& 47.2 & 39.8 
& 32.5 & 23.3 
& 36.1 & 27.3 
& 33.8 & 24.9 \\
Qwen-VL-Max &43.0&36.3&49.8&42.1&46.7&38.3&53.9&51.0&38.6&15.2&42.7&33.2&29.6&27.8\\
GPT-4o-mini  &51.3&44.5&58.7&54.8&58.2&53.2&59.6&55.2&41.1&26.0&57.4&50.1&49.7&47.6\\
Claude-3.5-Sonnet &56.9&49.6&60.8&52.6&58.7&50.3&64.0&58.3&45.0&25.4&56.5&48.0&48.1&45.2\\
GPT-4o       &55.1&47.8&61.0&56.8&58.7&54.4&62.2&58.2&45.2&30.0&58.6&52.6&54.1&51.0\\
GPT-4V       &56.2&53.4&69.1&57.1&65.0&55.0&72.0&61.4&48.1&30.3&61.8&46.3&42.0&36.7\\
Gemini-2.5-pro&62.0&55.3&68.3&61.9&66.1&60.8&68.9&64.1&52.1&37.1&65.7&57.9&57.0&54.6\\
GPT-5        &  {65.8}&  {60.1}&  {74.3}&  {68.1}&  {73.5}&  {66.7}&  {77.0}&  {69.2}&  {53.8}&  {44.7}&  {67.1}&61.7&60.3&  {57.5}\\ \rowcolor{backblue!75}
MathFlow$^{\star}_{\text{GPT\!-\!5}}$
&\textbf{66.5}&\textbf{61.8}&\textbf{74.6}&\textbf{68.5}&\textbf{73.8}&\textbf{67.2}&\textbf{77.0}&\textbf{69.3}&\textbf{58.2}&\textbf{54.1}&\textbf{70.2}&\textbf{66.7}&\textbf{69.4}&\textbf{66.2}\\
\bottomrule
\end{tabular}
\end{adjustbox}
\label{tab:full_flowverse_benchmark}
\vspace{-1em}
\end{table*}

\begin{table}[h]
   \centering
   \caption{\textbf{Full Ablation Analysis of MathFlow on FlowVerse$^{\dagger}$.}
   \textbf{Bold} numbers indicate the best performance.}
   \label{tab:sup_ablation}
   \vspace{-1em}
      \small
      \setlength{\tabcolsep}{3pt}
      \resizebox{1.\linewidth}{!}{
      \begin{tabular}{c c c|c}
         \toprule \textbf{\makecell[c]{Perception \\ Model for EI}} & \textbf{\makecell[c]{Perception \\ Model for RP}} & \textbf{\makecell[c]{Inference\\ Model}} & \textbf{\makecell[c]{COT-E\\(\%)}} \\
         \midrule Qwen2-VL-2B                                     & GPT-4V                                          & GPT-4V                                   & 49.0                               \\
         InternLM-XC2                                             & GPT-4V                                          & GPT-4V                                   & 47.1                               \\
         InfiMM-Math                                              & GPT-4V                                          & GPT-4V                                   & 52.6                               \\
         Qwen2-VL-7B                                              & GPT-4V                                          & GPT-4V                                   & 53.3                               \\
         InternVL2.5-8B                                           & GPT-4V                                          & GPT-4V                                   & 53.8                               \\
         Qwen2.5-VL-7B                                            & GPT-4V                                          & GPT-4V                                   & 53.9                               \\
         MathFlow-P-7B                                            & GPT-4V                                          & GPT-4V                                   & \textbf{58.9}                      \\
         \midrule GPT-4V                                          & Qwen2-VL-2B                                     & GPT-4V                                   & 49.8                               \\
         GPT-4V                                                   & InternLM-XC2                                    & GPT-4V                                   & 50.7                               \\
         GPT-4V                                                   & InfiMM-Math                                     & GPT-4V                                   & 54.1                               \\
         GPT-4V                                                   & Qwen2-VL-7B                                     & GPT-4V                                   & 52.9                               \\
         GPT-4V                                                   & MathFlow-P-7B                                   & GPT-4V                                   & \textbf{57.3}                      \\
         \midrule GPT-4V                                          & GPT-4V                                          & GPT-4V                                   & 56.2                               \\
         MathFlow-P-7B                                            & MathFlow-P-7B                                   & GPT-4V                                   & \textbf{59.3}                      \\
         \midrule MathFlow-P-7B                                   & MathFlow-P-7B                                   & GPT-4o-mini                              & {63.8}                             \\
         MathFlow-P-7B                                            & MathFlow-P-7B                                   & Llama3-80B                               & {67.9}                             \\
         MathFlow-P-7B                                            & MathFlow-P-7B                                   & Qwen2.5-72B                              & {69.6}                             \\
         MathFlow-P-7B                                            & MathFlow-P-7B                                   & GPT-4o                                   & {69.6}                             \\
         MathFlow-P-7B                                            & MathFlow-P-7B                                   & Gemini 2.5-pro                           & {70.4}                             \\
         MathFlow-P-7B                                            & MathFlow-P-7B                                   & DeepSeek-v3                              & {73.4}                      \\
         MathFlow-P-7B                                            & MathFlow-P-7B                                   & DeepSeek-r1                              & \textbf{75.6}                      \\
         \bottomrule
      \end{tabular}}
   \vspace{-0.5em}
\end{table}

\begin{table}[t]
   \centering
   \small
   \caption{\textbf{Full Performance Comparison of MLLMs on MathVerse and
   FlowVerse$^{\dagger}$ Datasets.} {FlowVerse}$^{\dagger}$ indicates the
   raw version of the dataset.
   }
   \label{tab:sup_reasoned}
   \vspace{-1em}
      \setlength{\tabcolsep}{3pt}
      \resizebox{1.\linewidth}{!}{
      \begin{tabular}{c c|c c}
         \toprule \textbf{\makecell[c]{Perception\\Model}} & \textbf{\makecell[c]{Inference\\Model}} & \textbf{MathVerse}           & \textbf{FlowVerse}$^{\dagger}$ \\
         \midrule \multicolumn{2}{c|}{InternLM-XC2}        & 25.9                                    & 39.6                          \\
         \multicolumn{2}{c|}{InfiMM-Math}                  & 34.5                                    & 47.1                          \\
         \multicolumn{2}{c|}{Qwen-VL-MaX}                  & 36.2                                    & 48.2                          \\
         \multicolumn{2}{c|}{Qwen2-VL-72B}                 & 38.9                                    & 52.3                          \\
         \multicolumn{2}{c|}{InternVL-2.5-78B}             & 43.2                                    & 54.7                          \\
         \multicolumn{2}{c|}{GPT-4V}                       & 54.4                                    & 56.2                          \\
         \multicolumn{2}{c|}{GPT-4o-mini}                  & 52.7                                    & 59.6                          \\
         \multicolumn{2}{c|}{GPT-4o}                       & 57.9                                    & 62.2                          \\
         \multicolumn{2}{c|}{Claude-sonnet-3.5}            & 57.4                                    & 64.0                          \\
         \multicolumn{2}{c|}{Gemini 2.5-pro}               & 59.9                                    & 68.9                          \\
         \midrule MathFlow-P-7B                            & InternLM-XC2                            & 30.2                         & 43.5                           \\
         MathFlow-P-7B                                     & InfiMM-Math                             & 38.1                         & 48.9                           \\
         MathFlow-P-7B                                     & Qwen-VL-MaX                             & 43.3                         & 54.3                           \\
         MathFlow-P-7B                                     & Qwen2-VL-72B                            & {48.1}                       & {58.3}                         \\
         MathFlow-P-7B                                     & InternVL-2.5-78B                        & \ul{56.8} & \ul{60.1}   \\
         MathFlow-P-7B                                     & GPT-4V                                  & {56.7}                       & {59.3}                         \\
         MathFlow-P-7B                                     & GPT-4o-mini                             & {58.9}                       & {63.8}                         \\
         MathFlow-P-7B                                     & GPT-4o                                  & {59.5}                       & {69.6}                         \\
         MathFlow-P-7B                                     & Claude-sonnet-3.5                       & {60.8}                       & {68.1}                         \\
         MathFlow-P-7B                                     & Gemini 2.5-pro                          & \textbf{62.4}  & \textbf{70.4}    \\
         \bottomrule
      \end{tabular}}
   \vspace{-0.5em}
\end{table}

\section{Additional Experimental Results}
\label{supsec:mathflow}
\subsection{More Analysis on MathVision}
\label{More_Analysis_of_Experiment_Results}
Tab.~\ref{tab:tab5} compares the performance of various closed and open-source Multimodal
Large Language Models (MLLMs) on the MathVision dataset across different mathematical
subjects. The highest scores are highlighted in bold for closed-source and uline
for open-source models.

MathFlow$^{\star}$$_{\text{GPT-V}}$ leads among closed-source models in
arithmetic (Ari), logical reasoning (Log), and geometry (Angle), showing its strength
in diverse mathematical domains. Among open-source models, MathFlow$^{\star}$$_{\text{Qwen2-VL-72B}}$
performs well in algebra (Alg), arithmetic (Ari), and topology (Topo),
demonstrating robust capabilities in visual mathematical tasks.

Overall, MathFlow models perform competitively with both closed and open-source baselines,
showcasing their adaptability to complex mathematical problem-solving by
effectively integrating visual perception information and logical reasoning.

\subsection{More Analysis of MathFlow Training}
\label{More_Detailed_Experiment_Results}
\textbf{Generate Config of MathFlow.} We first present the
generation configuration of MathFlow in Tab.~\ref{sup_Generate_Config}. These hyperparameters
were carefully tuned to optimize the model's generation performance while maintaining
output stability and reliability.

\vspace{0.3em}\noindent\textbf{Comprehensive and complete performance comparison across all models.}
We present comprehensive experimental results through three detailed
analyses. First, we evaluate our model's performance across different problem
versions in FlowVerse through Tab.~\ref{tab:full_flowverse_benchmark}. This
analysis spans six distinct versions incorporating various combinations of
textual and visual components: Descriptive Information (DI), Essential
Information (EI), Reasoned Property (RP), and Only Question (OQ). The Text Plus Version
specifically examines performance without image input. We report both standard
accuracy (Acc) and performance with the FlowVerse-CoT-E strategy (CoT-E), highlighting
the highest accuracies achieved by closed-source and open-source MLLMs in red and
blue respectively.

\vspace{0.3em}\noindent\textbf{Full Ablation Analysis of MathFlow on FlowVerse$^{\dagger}$.} Tab.~\ref{tab:sup_ablation} presents a thorough ablation analysis of our perception
model on FlowVerse$^{\dagger}$. This analysis systematically examines the contribution
of different components within our perception framework, with bold numbers
indicating the best performance across different configurations.

\vspace{0.3em}\noindent\textbf{Full Performance Comparison of MLLMs on
MathVerse and FlowVerse† Datasets.} Tab.~\ref{tab:sup_reasoned} provides a comprehensive performance
comparison of various MLLMs across both MathVerse and FlowVerse$^{\dagger}$ datasets.
Here, FlowVerse$^{\dagger}$ represents the raw version of the dataset, allowing us
to evaluate the models' fundamental capabilities without additional enhancements.
This comparison offers insights into the relative strengths of different architectures
and approaches across diverse mathematical reasoning tasks.

\begin{table}[t]
    \centering
      \small
      \caption{\textbf{Performance comparison under different thresholds.}}
      \vspace{-1em}
      \setlength{\tabcolsep}{8pt}
      \begin{tabular}{l|cccc}
        \toprule
        \textbf{Model} & \textbf{0.2} & \textbf{0.4} & \textbf{0.6} & \textbf{0.8} \\
        \midrule
        Qwen2-VL-72B    & 0.63 & 0.48 & 0.49 & 0.41 \\
        InfiMM-Math-7B  & 0.61 & 0.53 & 0.51 & 0.44 \\
        GPT-4V          & 0.44 & 0.36 & 0.40 & 0.32 \\
        Qwen2.5-VL-7B   & 0.37 & 0.31 & 0.29 & 0.25 \\
        \bottomrule
      \end{tabular}
      \label{tab:alpha_comparison}
        \vspace{-0.5em}
\end{table}

  \begin{table*}[t]
    \centering
      \small
      \caption{\textbf{The ablation study of the \textit{Image-Only} scenario.}}
      \vspace{-1em}
      \setlength{\tabcolsep}{8pt}
      \begin{tabular}{l|ccccc}
        \toprule \textbf{Model}              & \textbf{Text Centric$^{\dagger}$} & \textbf{Text Limited$^{\dagger}$} & \textbf{Vision Dense$^{\dagger}$} & \textbf{Vision Centric$^{\dagger}$} & \textbf{Vision Primary$^{\dagger}$} \\
        \midrule MathFlow$^{\star}_\text{Gpt-4o}$ & 68.5    (0.61 $\textcolor{darkgreen}{\downarrow}$)
        & 67.4 (0.80 $\textcolor{darkgreen}{\downarrow}$)& 52.8  (1.37 $\textcolor{darkgreen}{\downarrow}$)  & 55.4 (1.03 $\textcolor{darkgreen}{\downarrow}$) 
        & 54.7 (1.55 $\textcolor{darkgreen}{\downarrow}$)                        \\
        MathFlow$^{\star}_\text{DeepSeek-r1}$ & 72.1 (0.52 $\textcolor{darkgreen}{\downarrow}$)  & 70.4 (0.62 $\textcolor{darkgreen}{\downarrow}$)& 57.3 (0.97 $\textcolor{darkgreen}{\downarrow}$)                         & 65.7  (1.14 $\textcolor{darkgreen}{\downarrow}$)                           & 62.6  (1.39 $\textcolor{darkgreen}{\downarrow}$)                           \\
        \bottomrule
      \end{tabular}
                  \label{tab:ablationonly}
    \vspace{-0.5em}
  \end{table*}
  
Furthermore, as shown in Tab.~\ref{tab:alpha_comparison}, we also conducted a sensitivity analysis over different $\alpha$ values to guide the selection of $\alpha$. As shown in the table below, the overall model ranking and the relative performance gaps remain qualitatively stable across these settings, indicating that our main empirical conclusions do not depend on the particular choice of $\alpha$. At the same time, yields clear separation between models with different reasoning abilities without over-amplifying noise in either early or late steps, which motivates its use as our default setting.

\subsection{More Details of \textit{Image-Only} Scenario}
\label{Image-Only}
We introduce an \textit{Image-Only} variant (denoted by $\dagger$) in FlowVerse by superimposing each textual question onto its corresponding diagram, as illustrated in Tab.~\ref{tab:ablationonly}. We then evaluate our MathFlow pipeline on this variant. In the table below, green downward arrows (\textcolor{darkgreen}{$\downarrow$}) indicate the performance drop relative to the original multi-input setting. “MathFlow$^*_{4o}$” and “MathFlow$^*_{r1}$” denote MathFlow-P-7B paired with GPT-4o and DeepSeek-R1 inference models, respectively, and T and V abbreviate “Text” and “Vision.” These results confirm that MathFlow extends seamlessly to the \textit{Image-Only} scenario with only minor accuracy degradations.

\subsection{Reasoning Efficiency Analysis}
\label{Reasoning_Efficiency}

Tab.~\ref{tab:ablationRE} presents an evaluation of MathFlow built upon Qwen2.5-VL backbones ranging from 3B to 72B parameters, aiming to explore the trade-off between model size, reasoning accuracy, and inference latency on FlowVerse. As expected, both accuracy and inference time increase with model size—for instance, the 3B model achieves 34.2 accuracy in 0.47 seconds, while the 72B variant reaches 58.3 accuracy at the cost of 18.67 seconds per inference. The 7B variant strikes the most favorable balance, attaining 44.8 accuracy in just 2.51 seconds. Notably, the trained variant (MathFlow$_\text{train}$) achieves lower latency than the non-fine-tuned counterpart (MathFlow$_\text{ori}$), while avoiding redundant reasoning loops that occasionally occur in the latter.

\begin{table*}[t]
\centering
\small
\caption{\textbf{Reasoning efficiency of MathFlow built on Qwen2.5-VL models of varying sizes.} The trained variant MathFlowF$_\text{train}$) demonstrates faster inference than the non-fine-tuned version (MathFlow$_\text{ori}$), especially on the 7B model, which offers the best balance between speed and performance.}
\label{tab:ablationRE}
\vspace{-1em}
\setlength{\tabcolsep}{20pt}
\begin{tabular}{l|cccc}
\toprule \textbf{Model} & \textbf{Size=3B} & \textbf{Size=7B} & \textbf{Size=32B} & \textbf{Size=72B} \\
\midrule 
End2End  (CoT)          & 34.2 (0.47s) & 44.8 (2.51s) & 51.1 (6.45s)  & 58.3 (18.67s)  \\
MathFlow$_\text{ori}$ (CoT)       & 37.7 (0.68s) & 46.7 (3.46s) & 51.1 (9.13s)  & 59.2 (26.58s)  \\
MathFlow$_\text{train}$  (CoT)    & 48.0 (0.63s) & 51.7 (3.19s) & -            & -            \\
\bottomrule
\end{tabular}
\vspace{-0.5em}
\end{table*}

\subsection{Qualitative Examples}
\label{supsec:mathflow-Examples}

Figs.~\ref{re_case1}-\ref{re_case3} provide qualitative comparisons between GPT-4V
and MathFlow across several geometry problems. These examples illustrate common
error types, such as Visual Perception Errors and Reasoning Errors, which are
highlighted in the respective model responses.

In the first example, GPT-4V struggles with applying geometric properties correctly,
leading to a reasoning mistake while calculating the length of a chord. In contrast,
MathFlow arrives at the correct solution by following a systematic application
of the Pythagorean theorem and congruent relationships, demonstrating its advantage
in maintaining logical accuracy throughout the problem-solving process.

The second example highlights the challenge of dealing with circle theorems. GPT-4V
incorrectly deduces the angle at the center due to a misinterpretation of the
chord's properties, while MathFlow accurately follows through the geometric relationships
to determine the correct angle measure. This example underscores MathFlow’s
superior ability to navigate complex angle relationships and avoid error
propagation in multi-step reasoning.

The third example involves congruent triangles and corresponding parts. GPT-4V makes
a visual perception error by misidentifying the corresponding parts of the congruent
triangles, resulting in an incorrect response. MathFlow, on the other hand, correctly
identifies and matches corresponding sides and angles, showcasing its ability to
effectively manage both visual and logical components of geometric problems.

As illustrated in Figs.~\ref{re_case1}–\ref{re_case3}, end‐to‐end models often propagate perception errors (such as mislocalized points, incorrect angle estimations, or missed key annotations) into their final answers, resulting in entirely incorrect solutions. 
\textbf{MathFlow adopts a decoupled pipeline with two key advantages:}
1) By isolating the perception stage, MathFlow provides clear and inspectable intermediate outputs that make it straightforward to diagnose and correct vision‐based failures without retraining the entire reasoning model.
2) Once the raw visual content has been translated into structured textual representations, the downstream inference model is relieved of low‐level visual decoding tasks. It can apply its full capacity to logical and mathematical reasoning, thus reducing the complexity of the inference model and improving accuracy.

\begin{figure*}[b]
   \centering
   \includegraphics[width=1.0\textwidth]{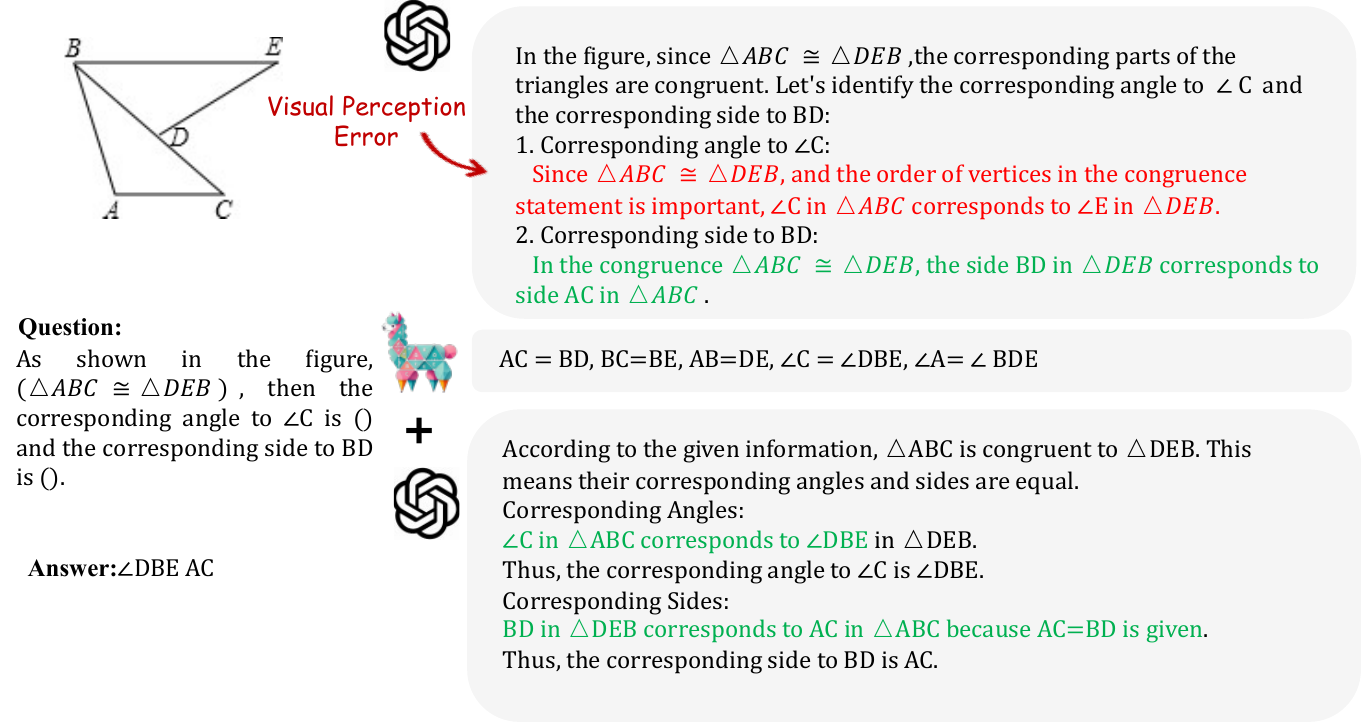}
   \vspace{-2em}
   \caption{\textbf{Response Comparison of GPT-4V and MathFlow$^{\star}$$_{\text{GPT-4V}}$}}
   \label{re_case1}
\end{figure*}

\begin{figure*}[b]
   \centering
   \includegraphics[width=1.0\textwidth]{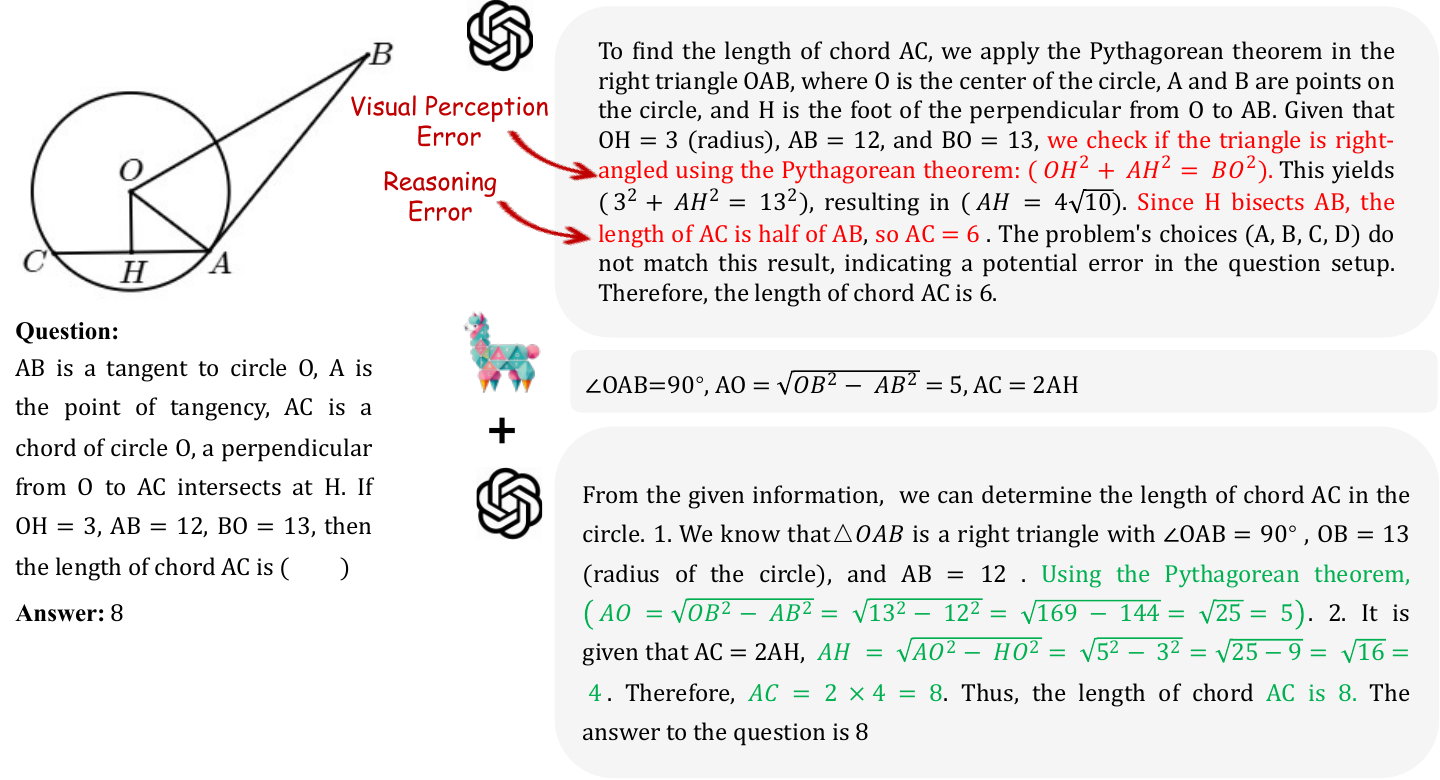}
   \vspace{-2em}
   \caption{\textbf{Response Comparison of GPT-4V and MathFlow$^{\star}$$_{\text{GPT-4V}}$}}
   \label{re_case2}
\end{figure*}

\begin{figure*}[b]
   \centering
   \includegraphics[width=1.0\textwidth]{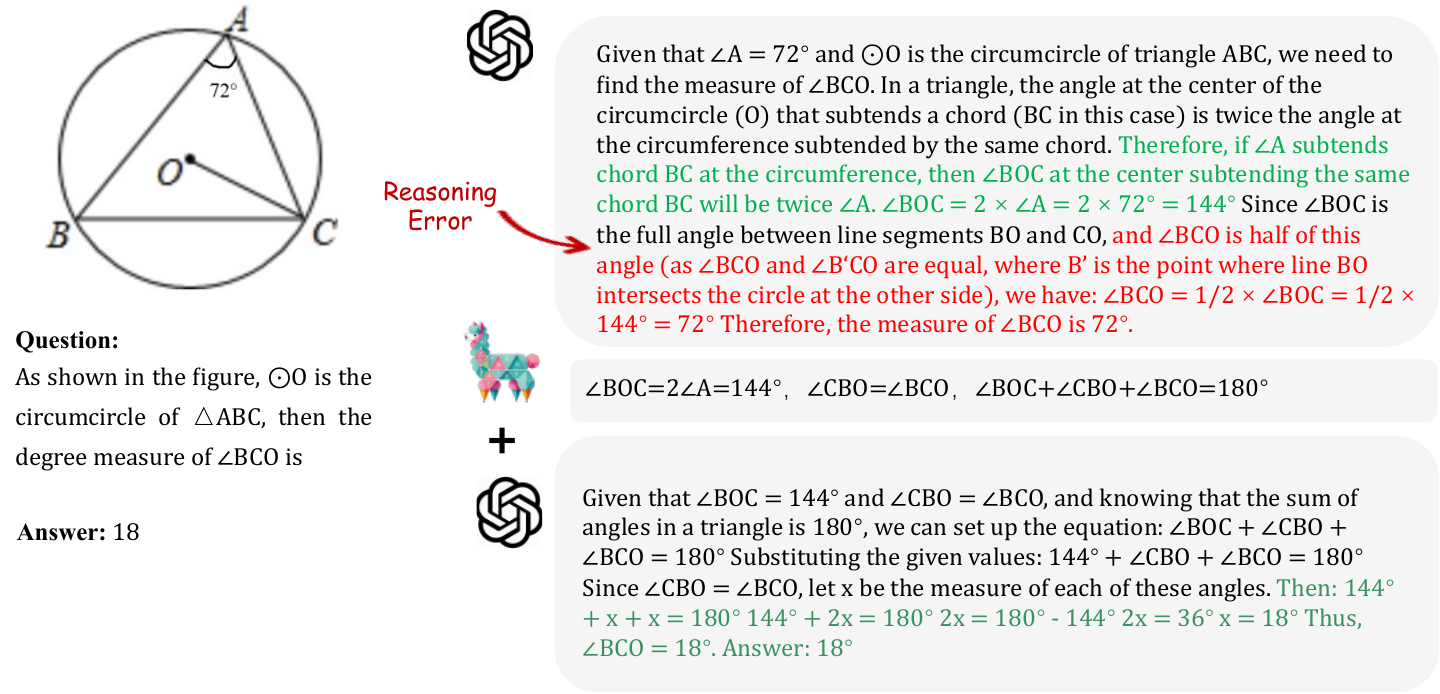}
   \vspace{-2em}
   \caption{\textbf{Response Comparison of GPT-4V and MathFlow$^{\star}$$_{\text{GPT-4V}}$}}
   \label{re_case3}
\end{figure*}

\end{document}